\documentclass[journal]{IEEEtran}
\ifCLASSINFOpdf
\else
\fi
\usepackage{times}  
\usepackage{helvet} 
\usepackage{courier}  
\usepackage[hyphens]{url}  
\usepackage{graphicx} 
\usepackage{caption} 

\usepackage{latexsym}
\usepackage{algorithm}
\usepackage{algorithmic}   
\usepackage{amsmath}
\usepackage{amsfonts}
\usepackage{url}
\usepackage[switch]{lineno}
\usepackage{booktabs}
\usepackage{threeparttable}  
\usepackage{graphicx}
\usepackage{subfigure}
\usepackage{multirow}
\usepackage{amssymb}
\usepackage{placeins}
\usepackage{color}
\usepackage{verbatim}
\usepackage{bm}
\usepackage{hyperref}
\begin{document}
%
\title{SpectralDiff: A Generative Framework for
Hyperspectral Image Classification with Diffusion Models
}
%
%
%

\author{Ning~Chen,
	Jun~Yue,
Leyuan~Fang,~\IEEEmembership{Senior~Member,~IEEE,}
	 and~Shaobo~Xia
	\thanks{This work was supported in part by the National Natural Science Foundation of China under Grant U22B2014, Grant 62101072 and Grant 42201481, in part by the Science and Technology Plan Project Fund of Hunan Province under Grant 2022RSC3064, in part by the Hunan Provincial Natural Science Foundation of China under Grant 2021JJ40570 and Grant 2023JJ40024. \emph{(Ning Chen and Jun Yue contributed equally to this work.) (Corresponding author: Leyuan Fang.)}}
	\thanks{Ning Chen is with the Institute of Remote Sensing and Geographic Information System, Peking University, Beijing 100871, China (e-mail: chenning0115@pku.edu.cn).}
	\thanks{Jun Yue is with the School of Automation, Central South University, Changsha 410083, China (e-mail: jyue@pku.edu.cn).}
	\thanks{Leyuan Fang is with the College of Electrical and Information Engineering, Hunan University, Changsha 410082, China, and also with the Peng Cheng Laboratory, Shenzhen 518000, China (e-mail: fangleyuan@gmail.com).}%
	\thanks{Shaobo Xia is with the Department of Geomatics Engineering, Changsha University of Science and Technology, Changsha 410114, China (e-mail: shaobo.xia@csust.edu.cn).}
	\thanks{}}

%
%

\markboth{}%
{Shell \MakeLowercase{\textit{et al.}}: Bare Demo of IEEEtran.cls for IEEE Journals}
%



\maketitle

\begin{abstract}
Hyperspectral Image (HSI) classification is an important issue in remote sensing field with extensive applications in earth science. In recent years, a large number of deep learning-based HSI classification methods have been proposed. However, existing methods have limited ability to handle high-dimensional, highly redundant, and complex data, making it challenging to capture the spectral-spatial distributions of data and relationships between samples. To address this issue, we propose a generative framework for HSI classification with diffusion models (SpectralDiff) that effectively mines the distribution information of high-dimensional and highly redundant data by iteratively denoising and explicitly constructing the data generation process, thus better reflecting the relationships between samples. The framework consists of a spectral-spatial diffusion module, and an attention-based classification module. The spectral-spatial diffusion module adopts forward and reverse spectral-spatial diffusion processes to achieve adaptive construction of sample relationships without requiring prior knowledge of graphical structure or neighborhood information. It captures spectral-spatial distribution and contextual information of objects in HSI and mines unsupervised spectral-spatial diffusion features within the reverse diffusion process. Finally, these features are fed into the attention-based classification module for per-pixel classification. The diffusion features can facilitate cross-sample perception via reconstruction distribution, leading to improved classification performance. Experiments on three public HSI datasets demonstrate that the proposed method can achieve better performance than state-of-the-art methods. For the sake of reproducibility, the source code of SpectralDiff will be publicly available at \href{https://github.com/chenning0115/SpectralDiff}{https://github.com/chenning0115/SpectralDiff}.
\end{abstract}


%
\IEEEpeerreviewmaketitle

\section{INTRODUCTION}
%
%
%
%

\IEEEPARstart{H}{yperspectral} imaging is a cutting-edge technology that enables the acquisition of high-resolution spectral information of objects. By integrating spatial and spectral reflectance information, each pixel in a hyperspectral image (HSI) corresponds to a unique spectral curve, providing rich information for identification and differentiation of diverse materials and surfaces. Beyond the limited perception of human eyes, the spectral detection range of hyperspectral imaging empowers a comprehensive understanding of nature \cite{8697135,7486259}. This technology has demonstrated significant potential in numerous fields, such as environmental management, agriculture, land management, ecology, geology, urban planning, and oceanography. HSI classification, which involves assigning pixels to specific land cover classes, such as soil and grass, is one of the most significant applications of hyperspectral imaging. As a fundamental component of hyperspectral data processing, HSI classification plays an indispensable role in most hyperspectral imaging applications \cite{8113128,7882742}. 

The high dimensionality of HSI poses a significant challenge for the accurate classification of pixels. With hundreds of spectral bands and massive amounts of data, it can be difficult to identify relevant features. To overcome this challenge, researchers have developed a range of methods to map spectral vectors from high-dimensional space to low-dimensional feature space in order to extract effective spectral features. These methods include classic statistical transformation techniques such as principal component analysis (PCA) \cite{4656486}, minimum noise fraction (MNF) \cite{7738467}, local preserving projection (LPP) \cite{7478572}, linear discriminant analysis (LDA) \cite{9625767}, independent component analysis (ICA) \cite{1634722}, and sparse preserving projection (SPP) \cite{xie2018low}. However, the spatial heterogeneity and homogeneity of HSIs make it difficult to fully utilize them by extracting spectral features alone. To address this limitation, researchers have proposed a series of methods to jointly extract spatial and spectral features, such as extended morphological profile (EMP) \cite{1396320} and extended attribute profile (EAP) \cite{6243172}. 

With the successful introduction and rapid development of deep neural networks (DNN) \cite{hinton2006reducing,9751593}, it has achieved outstanding performance in image classification \cite{He_2016_CVPR,wang2020large}, image segmentation \cite{7913730,9154595,10005033query}, instance segmentation \cite{10029885}, image vectorization \cite{10194945}, and object detection \cite{7112511, 7485869}. DNN offers a solution by offering an adaptable and potent framework for automatically learning complicated features and relationships in data \cite{9889110}. The HSI classification accuracy has been steadily improved by a number of HSI spectral feature extraction techniques based on DNN. Due to the fact that HSI has both spectral and spatial features, some spectral-spatial feature extraction techniques combining DNN have been proposed in order to fully explore the three-dimensional data characteristics \cite{9491800}, including stacked auto-encoders \cite{6844831}, deep fully convolutional network \cite{9491800}, deep prototypical network \cite{9170766}, spatial pyramid pooling \cite{yue2016deep}, and spectralformer \cite{9627165}.

Despite achieving favorable outcomes in HSI classification, DNN-based methods remain limited in their ability to model spectral-spatial relationships across samples. Current approaches primarily rely on graph neural networks (GNNs) for modeling sample relationships \cite{8474300, 9099071, 9170817, ssgrn}. However, using GNNs to measure the relationships between all samples requires designing graph structures or neighborhood information, which increases the complexity of the design and implementation processes and introduces subjectivity. From a data distribution standpoint, GNNs-based methods are ineffective in capturing the spectral-spatial distribution of data and do not fully represent the data generation process. As a result, these methods have limited perceptiveness towards contextual features when constructing relationships between samples.

To address the aforementioned challenge, we propose a generative framework based on diffusion models, named SpectralDiff. The proposed framework constructs the data generation process through iterative denoising, thereby obtaining spectral-spatial distribution information of the data and better capturing the relationships between samples. The framework consists of two modules, namely the spectral-spatial diffusion module and the attention-based classification module. In the spectral-spatial diffusion module, we use hyperspectral cube data with spatial and spectral dimensions and construct a hyperspectral channel distribution in the spectral-spatial latent feature space through the spectral-spatial diffusion process, which is a Markov process composed of forward and reverse processes. In the forward process, Gaussian noise is added to the hyperspectral channel, while in the reverse process, the noise is removed through multiple time steps with a spectral-spatial denoising network. By constructing the distribution of samples and the explicit sample generation process, the relationships between samples are constructed through the hidden variables of the spectral-spatial denoising network. We extract the spectral-spatial diffusion features that aggregate these latent variables and feed the generated features into the attention-based classification module. This module directly generates per-pixel classification results for hyperspectral data, thus achieving cross-sample perception and improving classification performance.

The primary innovative aspect of our study is adopting a generative perspective, wherein we demonstrate the process of modeling sample generation to acquire spectral-spatial features infused with contextual information of the spectral-spatial distribution. This is achieved without the need for pre-defined graph structures or neighborhood information. It is worth noting that our proposed generative framework is loosely coupled between each module, which can evolve independently in the future, thus boosting the development of HSI classification. The main contributions of this paper can be summarized as follows.

\begin{itemize}
\item We formulate the construction of relationships between samples from a generative perspective, designated as a spectral-spatial diffusion process. To the best of our knowledge, this is the first study to apply diffusion models to HSI classification.
\item We present an HSI classification framework based on forward and reverse diffusion processes, which harnesses the iterative generative process of constructing samples for obtaining spectral-spatial features endowed with contextual information, all without the need for pre-defined graph structures or neighborhood information.
\item Numerous experiments demonstrate the superiority of the proposed method over existing state-of-the-art techniques. Moreover, through ablation experiments, we validate the effectiveness of spectral-spatial diffusion features.
\end{itemize}

The rest of this article is structured as follows. In Section II, we provide a brief overview of the existing work on HSI classification and diffusion models. Section III outlines our proposed SpectralDiff in detail, highlighting the key components of our approach. In Section IV, we present the experimental setup and discuss our results from both qualitative and quantitative perspectives. Additionally, we validate the effectiveness of our approach through ablation experiments. Finally, in Section V, we summarize the conclusions of our study and provide insights into the future directions of research in this field.

\section{Related Work}

In this section, we provide an introduction to the background and related work associated with our proposed method. Specifically, we discuss existing techniques for HSI classification based on traditional spectral-spatial feature extraction, as well as those based on deep learning. Furthermore, we present an overview of the diffusion model and its development context, highlighting the key advancements and challenges in this field. Through this exposition, we aim to establish a solid foundation for our proposed method and contextualize our contribution within the broader scope of HSI classification research.

\subsection{Hyperspectral Image Classification}
Hyperspectral sensor captures both spatial and spectral information. In an HSI image, every pixel vector corresponds to a unique spectral curve. The primary objective of HSI image classification is to assign each pixel to a specific land cover class, such as river, forest, lake, farmland, building, grassland, mineral, road and rock. Classification is a pivotal step in the application of HSI and has significant implications for environment, geology, mining, ecology, forestry, agriculture, and other areas \cite{6297992,7882742}. HSI classification enables quick and precise acquisition of ground feature, empowering informed decision-making and management. For instance, in agriculture, HSI classification can be used to track disease progression and crop development, improving crop quality and productivity. In the mining sector, HSI classification can make it easier to find and identify minerals, increasing the effectiveness of mining natural resources \cite{8113128,9186822}.

Spectral feature extraction based on continuous spectral signals is widely used in HSI feature extraction, which considers each pixel of the HSI as an independent spectral vector and ignores the spatial relationships between pixels when generating the ground objects' features \cite{9360315}. Linear feature extraction methods such as principal component analysis (PCA) \cite{4656486}, linear discriminant analysis (LDA) \cite{9625767}, and minimum noise fraction (MNF) \cite{7738467, 9846880} are among the commonly used spectral feature extraction techniques. Nonetheless, the nonlinear nature of HSI has led to the development of numerous nonlinear feature extraction approaches in recent years, complementing the statistical transformation methods based on prior knowledge. In recent years, many nonlinear HSI feature extraction methods for have been proposed \cite{9082155,8697135}, including local Fisher discriminant analysis (LFDA) \cite{5765424}, manifold learning \cite{8677267}, sparsity preserving projections (SPP) \cite{6587316}, improved manifold coordinate representations \cite{1704966} and locality preserving projections (LPP) \cite{7478572}.

The HSI analysis technique involves extracting separable features of patterns from the spectral signal to reduce the dimensionality of the data. Despite its efficacy, this method alone cannot fully utilize the characteristics of HSI due to its spatial homogeneity and heterogeneity. Consequently, incorporating image texture and structural features is necessary to overcome this limitation. Researchers have proposed several methods to extract spatial structure and texture information from HSIs to obtain spatial features such as three-dimensional gray-level cooccurrence \cite{6410025} and discriminative Gabor feature selection \cite{6194995}. In the domain of HSI feature extraction and classification, the classic spectral-spatial joint feature extraction methods include extended morphological profile \cite{1396320}, spatial and spectral regularized local discriminant embedding \cite{6856200}, extended attribute profile \cite{6243172}, spatial–spectral manifold alignment \cite{9256351}, directional morphological profiles \cite{1396320}, and spectral-spatial locality preserving projection \cite{kianisarkaleh2016spatial}.

Deep learning has emerged as a significant breakthrough in the field of machine learning, providing automatic feature learning from data \cite{hinton2006reducing}. HSI classification, in particular, benefits from deep learning models due to their ability to handle complex and nonlinear relationships between input data and output classes \cite{wang2020large}, leading to improved classification accuracy compared to traditional machine learning methods \cite{yue2015spectral,9846880}. To fully utilize the spectral-spatial features of HSI, joint spectral-spatial HSI feature extraction methods based on deep learning, such as recurrent neural network \cite{7914752}, deep residual network \cite{8061020,8445697}, capsule networks \cite{8509610}, and transformer \cite{9627165}, have been proposed. While these methods have yielded satisfactory results, challenges remain, such as the inability to capture global relationships between samples. To address this issue, researchers have proposed using graph neural network to model the relationship between samples \cite{8474300}. However, the use of graph neural network for global relationship modeling incurs high computing and memory costs, and slow gradient decline \cite{9170817}. In this paper, we propose using diffusion models to model the global relationship of samples and verify the effectiveness of this method through extensive comparative experiments.

\subsection{Diffusion Models}
Diffusion models, also known as diffusion probabilistic models \cite{nichol2021improved}, belong to the class of latent variable models (LVM) in machine learning \cite{kingma2021variational}. They draw inspiration from non-equilibrium thermodynamics and construct a Markov chain that gradually introduces random noise into the input data \cite{sohl2015deep}. Subsequently, the Markov chain is trained utilizing variational inference \cite{croitoru2022diffusion}, delivering remarkable performance across various domains such as natural language processing \cite{NEURIPS2022_ec795aea,10149431}, time series forecasting \cite{alcaraz2022diffusion,rasul2021autoregressive}, and molecular graph modeling \cite{zhang2023survey,jing2022torsional}. Currently, research on diffusion models is generally based on three primary paradigms \cite{yang2022diffusion}, namely score-based generative models \cite{song2019generative,NEURIPS2021_5dca4c6b,NEURIPS2020_92c3b916}, stochastic differential equations (SDE) \cite{song2020score}, and denoising diffusion probabilistic models (DDPM) \cite{NEURIPS2020_4c5bcfec,nichol2021improved}.

The diffusion model aims to infer the underlying structure of a dataset by modeling the diffusion of data points in the latent space \cite{dhariwal2021diffusion}, which consists of two core processes: forward and reverse. During the forward process, the input data is gradually perturbed by introducing Gaussian noise in multiple time steps. Conversely, the reverse process aims to restore the original input data by reducing the discrepancy between the predicted noise and the increased noise in multiple reverse time steps. In the field of computer vision, this entails training a neural network to perform denoising of images that are blurred by Gaussian noise, by learning the reverse diffusion process \cite{yue2023dif}. The diffusion model has gained popularity in recent times, owing to its remarkable flexibility and robustness, and has been successfully employed to address a diverse range of intricate visual challenges \cite{croitoru2022diffusion}, such as image inpainting \cite{lugmayr2022repaint}, image generation \cite{sohl2015deep,NEURIPS2020_4c5bcfec,song2020score,dhariwal2021diffusion}, image-to-image translation \cite{saharia2022palette,zhao2022egsde}, image fusion \cite{yue2023dif}, and image super-resolution \cite{9887996,daniels2021score,chung2022come}.

The feature representation learned from the diffusion models has been demonstrated to be highly effective in various discriminating tasks, such as image classification \cite{zimmermann2021score}, object detection \cite{chen2022diffusiondet} and image segmentation \cite{baranchuk2021label,amit2021segdiff}. By constructing the distribution of input data, the diffusion models can effectively capture the underlying patterns and relationships between samples. In this paper, we propose a generative framework for hyperspectral classification based on the diffusion models, which explicitly constructs the data generation process through hierarchical denoising. The proposed framework is able to build the distribution of multi-channel input data and effectively exploit the distributional information of high-dimensional and highly redundant data, thereby improving the performance of HSI classification.

\begin{figure*}[t]
	\centering 
	{\tiny }	\includegraphics[width=\textwidth]{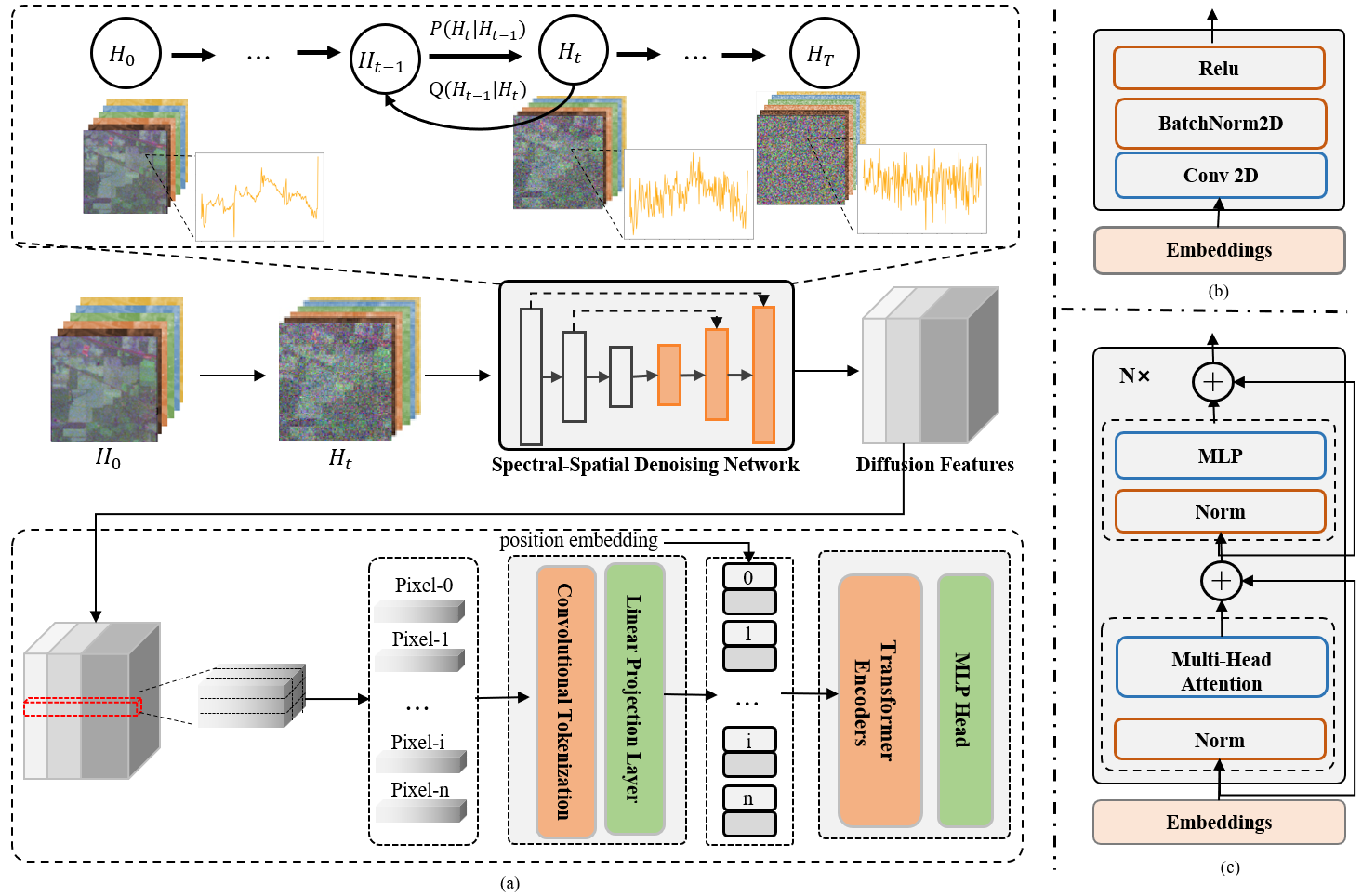} 
	\caption{Overview of the proposed SpectralDiff. (a) The architecture of the generative framework with diffusion process. $\bm{\mathcal{H}_{0}}$ and $\bm{\mathcal{H}_{t}}$ represent hyperspectral images of timestep $0$ and timestep $T$, respectively. $P(\cdot|\cdot)$ and $Q(\cdot |\cdot)$ represent the forward and reverse spectral-spatial diffusion processes, respectively. (b) Structure of Convolutional Tokenization. (c) Structure of Transformer Encoders.
 } 
	\label{overall} 
\end{figure*}

\section{METHODOLOGY}
In this section, we provide a detailed description of the proposed SpectralDiff, as shown in Fig.~\ref{overall}. By training the spectral-spatial denoising network to estimate the noise added in the forward spectral-spatial diffusion process, we establish the relationship among all samples. 

\subsection{Spectral-Spatial Diffusion Module}
Given an HSI $\bm{\mathcal{H}} \in \mathbb{R}^{H\times W\times B}$, where $H$ and $W$ denote the height and width of $\bm{\mathcal{H}}$, respectively. $B$ represents the number of spectral channels. 
To learn the joint latent structure of images with hundreds of channels, we add noise to the spectral-spatial instance in the forward process, and eliminate the noise added by the forward process by training a spectral-spatial denoising network in the reverse process \cite{NEURIPS2020_4c5bcfec}. The purpose of training the diffusion model with forward and reverse processes is to learn the joint latent structure between hyperspectral channels by simulating the diffusion of hyperspectral channels in the latent space \cite{NEURIPS2020_92c3b916,Gu_2022_CVPR}.
\begin{figure*}[!h]
	\centering 
	\includegraphics[width=0.99\textwidth]{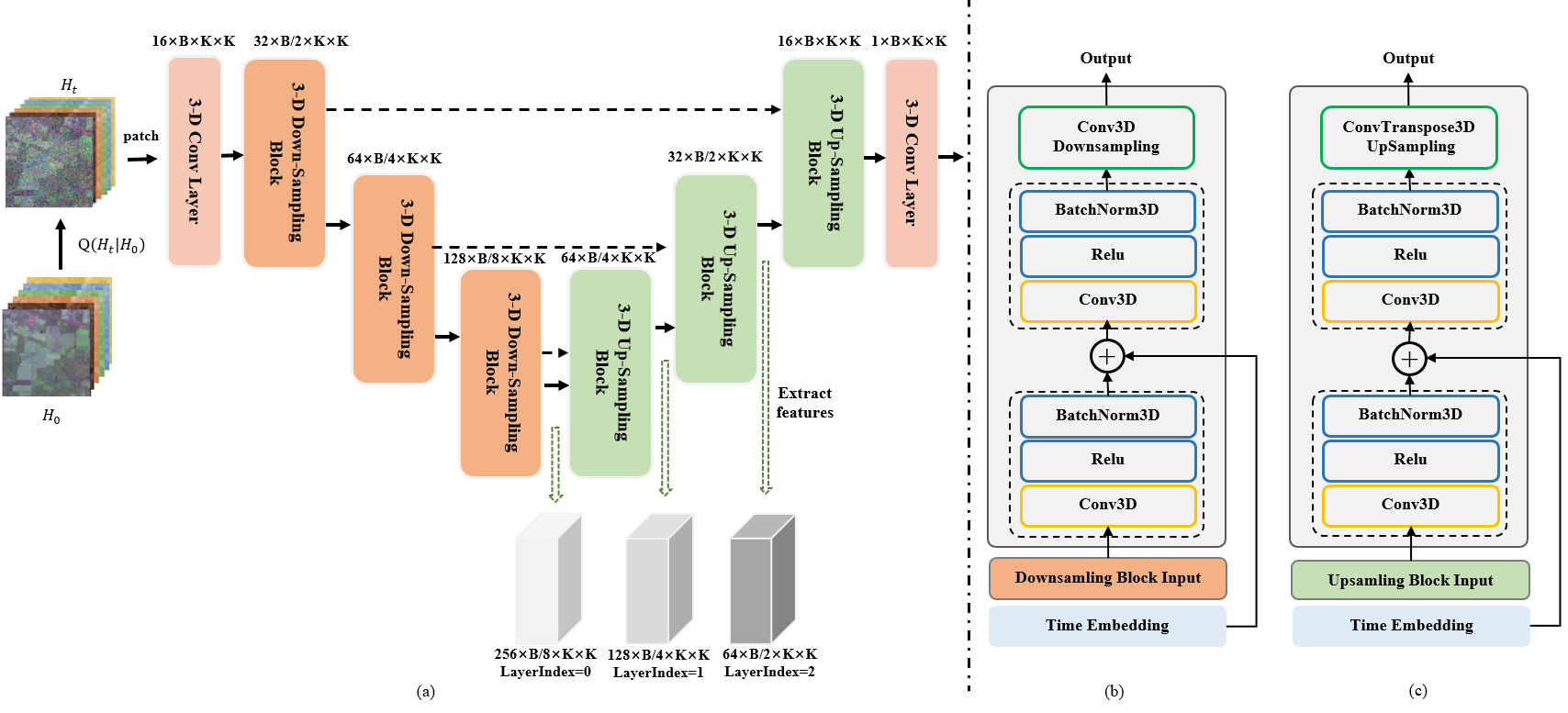} 
	\caption{Structure of the Spectral-Spatial Denoising Network. (a) The original HSI data is processed through a forward diffusion process to obtain image data with added Gaussian noise. Subsequently, the data is passed through an encoder-decoder structure and finally output as a tensor with the same shape as the original HSI data. This output represents the network's prediction for the initial noise $\epsilon$ added. (b) The details of the 3-D down-sampling block. (c) The details of the 3-D up-sampling block.} 
	\label{fig:unet} 
\end{figure*}
\subsubsection{Forward Spectral-Spatial Diffusion Process}
The forward spectral-spatial diffusion process, drawing inspiration from the principles of non-equilibrium thermodynamics, can be considered as Markov chain \cite{sohl2015deep,NEURIPS2020_4c5bcfec}. Its progressive stages entail the gradual incorporation of Gaussian noise into the data. In this study, we extract spectral-spatial instance $\bm{\mathcal{HI}} \in \mathbb{R}^{K\times K\times B}$ using a $K\times K$ neighborhood from the whole HSI $\bm{\mathcal{H}}$. The forward spectral-spatial diffusion process is distinguished by the ``no memory'' property, whereby the probability distribution of the HSI at a given time $t+1$ is exclusively determined by its state at time $t$. At the $t$th step, the spectral-spatial instance imbued with noise is expressed as follows:
\begin{eqnarray}
	P(\bm{\mathcal{HI}_{t}}|\bm{\mathcal{HI}_{t-1}})=\mathcal{N}(\bm{\mathcal{HI}_{t}};\sqrt{\alpha_{t}}\bm{\mathcal{HI}_{t-1}},(1-\alpha_{t})\bm{I})
	\label{xt}
\end{eqnarray}
where $\bm{\mathcal{HI}_{t-1}}$ and $\bm{\mathcal{HI}_{t}}$ denote the noisy hyperspectral instances at timestep $t-1$ and $t$, respectively. $\bm{I}$ stands for the standard normal distribution. $\sqrt{\alpha_{t}}\bm{\mathcal{HI}_{t-1}}$ and $(1-\alpha_{t})\bm{I}$ denote the mean and variance of $P(\bm{\mathcal{HI}_{t}}|\bm{\mathcal{HI}_{t-1}})$, respectively. The variance of Gaussian noise added at timestep $t$ is controlled by a variance schedule referred to as $\alpha_{t}$. The magnitude of the added Gaussian noise decreases as the value of $\alpha_{t}$ increases. By Eq.~\eqref{xt}, we can derive the expression of $\bm{\mathcal{HI}_{1}} \in \mathbb{R}^{K\times K\times B}$ during the first diffusion as follows:
\begin{eqnarray}
	\bm{\mathcal{HI}_{1}}=\sqrt{\alpha_{1}}\bm{\mathcal{HI}_{0}}+\sqrt{1-\alpha_{1}}\bm{\epsilon}
	\label{x1}
\end{eqnarray}
where $\bm{\mathcal{HI}_{0}}$ stands for the hyperspectral instance before diffusion. $\bm{\epsilon} \in \mathbb{R}^{K\times K\times B}$ is the added Gaussian noise. By using Eq.~\eqref{xt} and Eq.~\eqref{x1}, the expression of $\bm{\mathcal{HI}_{t}}$ can be derived as follows:
\begin{eqnarray}
\left\{
\begin{aligned}
&\bm{\mathcal{HI}_{t}}=\sqrt{\bar{\alpha}_{t}}\bm{\mathcal{HI}_{0}}+\sqrt{1-\bar{\alpha}_{t}}\bm{\epsilon}\\
&\bar{\alpha}_{t}=\prod \limits_{i=1}^t\alpha_{i}
\end{aligned}
\label{x0xt}
\right.
\end{eqnarray}
where $\bar{\alpha}_{t}$ represents the product of $\alpha_{1}$ to $\alpha_{t}$.
The computation of $\bm{\mathcal{HI}_{t}}$ in the context of forward spectral-spatial diffusion process hinges upon the timestep $t$, variance schedule $\alpha_{1},...,\alpha_{t}$, and the noise sampled from the standard normal distribution. Given these inputs, the hyperspectral instance at timestep $t$ can be directly generated by Eq.~\eqref{x0xt}.

\subsubsection{Reverse Spectral-Spatial Diffusion Process}
In the process of reverse spectral-spatial diffusion, a spectral-spatial denoising network is trained to gradually denoise the noisy hyperspectral instance to obtain the original hyperspectral instance $\bm{\mathcal{HI}_{0}}$ \cite{baranchuk2021label}. In the $t$-th step of the reverse diffusion process, denoising operation is performed on the noisy hyperspectral instance $\bm{\mathcal{HI}_{t}}$ to obtain the hyperspectral instance of the previous step, which is $\bm{\mathcal{HI}_{t-1}}$. Given the hyperspectral instance at step $t$, the conditional probability of the hyperspectral instance at step $t-1$ can be expressed as follows \cite{NEURIPS2020_4c5bcfec,gedara2022remote}:
\begin{eqnarray}
	Q(\bm{\mathcal{HI}_{t-1}}|\bm{\mathcal{HI}_{t}})=\mathcal{N}(\bm{\mathcal{HI}_{t-1}};\mu_{\theta}(\bm{\mathcal{HI}_{t}},t),\sigma_{t}^{2}\bm{I})
	\label{xtxt-1}
\end{eqnarray}
where $\mu_{\theta}(\bm{\mathcal{HI}_{t}},t)$ and $\sigma_{t}^{2}$ denote the mean and variance of the condition distribution $Q(\bm{\mathcal{HI}_{t-1}}|\bm{\mathcal{HI}_{t}})$, respectively. The variance is determined by the variance schedule, which can be expressed as follows:
\begin{eqnarray}
	\sigma_{t}^{2}=\frac{1-\bar{\alpha}_{t-1}}{1-\bar{\alpha}_{t}}(1-\alpha_{t})
	\label{variance}
\end{eqnarray}

To obtain the mean of the conditional distribution $Q(\bm{\mathcal{HI}_{t-1}}|\bm{\mathcal{HI}_{t}})$, we need to train the network to predict the added noise. The mean can be represented as follows:
\begin{eqnarray}
	\mu_{\theta}(\bm{\mathcal{HI}_{t}},t)=\frac{1}{\sqrt{\alpha_{t}}}(\bm{\mathcal{HI}_{t}}-\frac{1-\alpha_{t}}{\sqrt{1-\bar{\alpha}_{t}}}\epsilon_{\theta}(\bm{\mathcal{HI}_{t}},t))
	\label{mean}
\end{eqnarray}
where $\epsilon_{\theta}(\cdot,\cdot)$ denote the spectral-spatial denoising network whose input is the timestep $t$ and the noisy hyperspectral instance $\bm{\mathcal{HI}_{t}}$ at timestep $t$.

\subsection{Loss Function of Spectral-Spatial Diffusion Process}
For the spectral-spatial denoising network training, the hyperspectral instance $\bm{\mathcal{HI}_{0}^{i}} \in \mathbb{R}^{K\times K\times B}$ are first extracted, followed by sampling the noise matrix $\bm{\epsilon^{i}} \in \mathbb{R}^{K\times K\times B}$ of equivalent size from the standard normal distribution. Subsequently, the timestep $t$ is sampled from a uniform distribution $U(\{1,...,T\})$. With the above-mentioned sampling completed, the noisy hyperspectral instance at timestep $t$ can be calculated by Eq.~\eqref{x0xt}. The noisy hyperspectral instance and timestep are fed into the spectral-spatial denoising network to generate the predicted noise. The loss function of the spectral-spatial diffusion process can be expressed as follows:
\begin{eqnarray}
\begin{aligned}
	\mathcal{L}_{ssdp} &=\sum_{i=1}^N\left \| \bm{\epsilon^{i}}-\epsilon_{\theta}(\bm{\mathcal{HI}_{t}^{i}},t)\right \|_{1}\\
 &=\sum_{i=1}^N\left \| \bm{\epsilon^{i}}-\epsilon_{\theta}(\sqrt{\bar{\alpha}_{t}}\bm{\mathcal{HI}_{0}^{i}}+\sqrt{1-\bar{\alpha}_{t}}\bm{\epsilon},t)\right \|_{1}
 \end{aligned}
	\label{loss}
\end{eqnarray}
where $N$ and $i$ represent the total number of hyperspectral instances and the index of a hyperspectral instance, $\epsilon_{\theta}(\cdot,\cdot)$ represents the spectral-spatial denoising network discussed in the previous section. This network operates by taking two input components: the time step $t$ and the corresponding hyperspectral image $\bm{\mathcal{HI}_{t}^{i}}$ with Gaussian noise added. The symbol $\|\cdot\|_{1}$ denotes the L1 norm. This loss function quantifies the discrepancy between the initial noise $\bm{\epsilon^{i}}$ and the estimated noise. Consequently, the spectral-spatial denoising network has the capability to predict the magnitude of noise introduced to the original image.

\subsection{Structure of the Spectral-Spatial Denoising Network}
The structure of the spectral-spatial denoising network adopts a similar architecture to U-Net \cite{9887996}, as shown in Fig.~\ref{fig:unet}. The network takes the time step $t$ and the corresponding hyperspectral image $\bm{\mathcal{H}_{t}}$ with added Gaussian noise as inputs. Considering the potentially large size of remote sensing images, we used the patch form of the image $\bm{\mathcal{HI}_{t}}$ as the input instead. The network outputs predictions for the Gaussian noise $\epsilon$. The loss function is shown in Eq.~\eqref{loss}, as described in previous section.

Specifically, the input undergoes one 3D convolution layer and three 3D down-sampling blocks to achieve spectral and spatial feature encoding. Subsequently, it is decoded via three 3D up-sampling blocks and another 3D convolution layer to generate the output. Unlike diffusion model processing RGB images, the Spectral-Spatial Denoising Network employs 3D convolution structures for the efficient extraction of spectral and spatial information simultaneously during down-sampling and up-sampling processes. To extract the diffusion feature of every pixel in both encoder and decoder processes, we preserved the spatial dimensions while compressing and restoring the spectral dimensions. The internal structures of 3D down-sampling and 3D up-sampling blocks are depicted in Fig.~\ref{fig:unet}(b) and Fig.~\ref{fig:unet}(c), respectively, consisting of two-layer blocks that incorporate a 3D-convolution layer, a BatchNorm3D layer, and a ReLU activation layer. The down-sampling structure applies a 3D convolution operation with a stride of 2 to gradually reduce the spectral dimension, while the up-sampling block utilizes a 3D deconvolutional layer with a stride of 2 to gradually restore the spectral dimension.

\subsection{Attention-based Classification Module}
Once the training of the spectral spatial denoising network is completed, we will employ the network to generate spectral-spatial diffusion features. Specifically, we utilize the high-hyperspectral image $\bm{\mathcal{H}_{t}}$ corrupted by Gaussian noise at time step $t$ as the network's input. During the network's inference process, we extract the activation tensors from the down-sampling and up-sampling blocks as diffusion features. As depicted in Fig.~\ref{fig:unet}(a), in practical implementation, the U-net's presence of shortcut connections enables us to directly obtain the input of each up-sampling layer for the corresponding feature extraction. Importantly, the obtained diffusion features may differ in information across different timestamps and up-sampling layers, potentially affecting the final classification performance.

Fig.~\ref{overall}(a) illustrates the fundamental architecture of the attention-based classification module. Here, we denote the extracted diffusion features as $F \in R^{H \times W \times L}$, where $H$, $W$, and $L$ represent height, width, and the feature channel, respectively. 
 It is worth noting that the spatial dimension of the diffusion features is consistent with that of the raw features. For each pixel $i$, the corresponding channel dimension not only contains the original spectral information, but also includes the deep features fused through the spectral-spatial diffusion module. we use the patch $Z_i \in R^{k \times k \times L}$, which is centered on the pixel, as the input to the classifier. The spectral-spatial features of $Z_i$ are mapped via Conv2D structure and linear mapping layer. This process produces a token sequence that is appropriate for processing with Transformer. During this process, the region surrounding pixel $i$ is separated into distinct spatial tokens, which are then consolidated through the attention mechanism. The kernel of the Conv2D is set to $3 \times 3$. To provide positional information, the model adds position embedding to the input information before feeding it to the transformer encoder. The transformer encoder structure is depicted in Fig.~\ref{overall}(b). It comprises multiple identical substructures, each of which contains a multi-head attention layer and an MLP layer. Finally, the model maps the output to the classification through the MLP layer. The attention-based classification module combines the CNN and transformer structures to form an effective classifier. This approach utilizes a CNN structure to perform spectral-spatial feature mapping, and a Transformer structure for deep feature extraction, which leads to exceptional classification performance.

\section{Experiments}
\subsection{Experimental Settings}
\subsubsection{Datasets}
\begin{table*}[htbp]
	\small 
	\begin{center}
		\caption{Training and Test Sample Numbers in the Indian Pines Dataset, the Pavia University Dataset, and the Salinas Dataset}
		\setlength{\tabcolsep}{2mm}
		\label{tab:DataSet}
\begin{tabular}{c|ccc|ccc|ccc}\toprule\multicolumn{1}{c|}{\multirow{2}{*}{No.}} &\multicolumn{3}{c|}{\textbf{Indian Pines}}&\multicolumn{3}{c|}{\textbf{Pavia University}}&\multicolumn{3}{c}{\textbf{Salinas}}\\ \multicolumn{1}{c|}{}  & \multicolumn{1}{c}{Class} & \multicolumn{1}{c}{Train.} & \multicolumn{1}{c|}{Test.} &\multicolumn{1}{c}{Class} & \multicolumn{1}{c}{Train.} & \multicolumn{1}{c|}{Test.} &\multicolumn{1}{c}{Class} & \multicolumn{1}{c}{Train.} & \multicolumn{1}{c}{Test.}  \\\midrule \midrule 1  &  Alfalfa  &  30  &  16  &  Asphalt  &  30  &  6601  &  BrocoliGreenWeeds1  &  30  &  1979  \\  2  &  CornMotill  &  30  &  1398  &  Meadows  &  30  &  18619  &  BrocoliGreenWeeds2  &  30  &  3696  \\  3  &  CornMintill  &  30  &  800  &  Gravel  &  30  &  2069  &  Fallow   &  30  &  1946  \\  4  &  Corn  &  30  &  207  &  Trees  &  30  &  3034  &  FallowRoughPlow  &  30  &  1364  \\  5  &  GrassPasture  &  30  &  453  &  PaintedMetalSheets  &  30  &  1315  &  FallowSmooth  &  30  &  2648  \\  6  &  GrassTrees  &  30  &  700  &  BareSoil  &  30  &  4999  &  Stubble  &  30  &  3929  \\  7  &  GrassPastureMowed  &  14  &  14  &  Bitumen  &  30  &  1300  &  Celery  &  30  &  3549  \\  8  &  HayWindrowed  &  30  &  448  &  SelfBlockingBricks  &  30  &  3652  &  GrapesUntrained  &  30  &  11241  \\  9  &  Oats  &  10  &  10  &  Shadows  &  30  &  917  &  SoilVinyardDevelop  &  30  &  6173  \\  10  &  SoybeanNotill  &  30  &  942  &     &     &     &  ComSenescedGreenWeeds  &  30  &  3248  \\  11  &  SoybeanMintill  &  30  &  2425  &     &     &     &  LettuceRomaine4wk  &  30  &  1038  \\  12  &  SoybeanClean  &  30  &  563  &     &     &     &  LettuceRomaine5wk  &  30  &  1897  \\  13  &  Wheat  &  30  &  175  &     &     &     &  LettuceRomaine6wk  &  30  &  886  \\  14  &  Woods  &  30  &  1235  &     &     &     &  LettuceRomaine7wk  &  30  &  1040  \\  15  &  BuildingsGrassTreesDrives  &  30  &  356  &     &     &     &  VinyardUntrained  &  30  &  7238  \\  16  &  StoneSteelTowers  &  30  &  63  &     &     &     &  VinyardVerticalTrellis  &  30  &  1777  \\
\midrule
\midrule
&  Total  &  444  &  9805  &  Total  &  270  &  42506  &  Total  &  480  &  53649 \\
\bottomrule \end{tabular}
	\end{center}
\end{table*}
In order to verify the effectiveness of our algorithm, we applied the algorithm to three public datasets: the Indian Pines dataset, the Pavia University dataset, and the Salinas dataset.

The Indian Pines (IP) dataset was collected in 1992 using the Airborne Visible/Infrared Imaging Spectrometer (AVIRIS) Sensor, covering the northwestern region of Indiana in the United States. Comprising of 224 spectral bands, the uncorrected dataset spans a range of 0.4 to 2.5$\mu$m and is composed of 145×145 pixels with a spatial resolution of 20 meters. It contains 16 different land-cover classes. For the purpose of experimentation, 24 water-absorption bands and noise bands were removed, and a subset of 200 bands were selected for our analysis. 

\begin{figure}[!h]
	\centering 
	\includegraphics[width=0.50\textwidth]{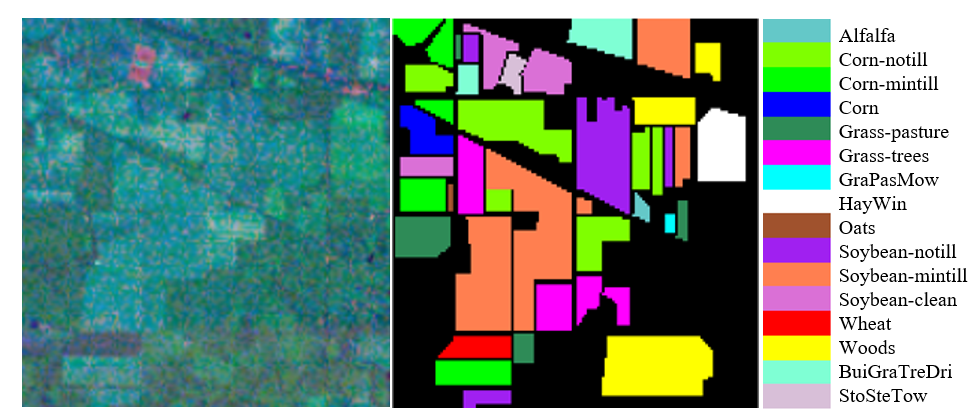} 
	\caption{The false-color composite image, the corresponding ground-truth map, and the legend of the Indian Pines dataset.} 
	\label{fig:indian_data} 
\end{figure}

The Pavia University (PU) dataset was collected in 2001 by the Reflective Optics System Imaging Spectrometer (ROSIS) Sensor, covering the Pavia University in Northern Italy. It contains of 115 spectral bands while 12 noise bands were removed so that 103 bands were used. It is composed of 610 × 340 pixels with a spatial resolution of 1.3 m. The dataset convers Nine categories.

\begin{figure}[!h]
	\centering 
	\includegraphics[width=0.5\textwidth]{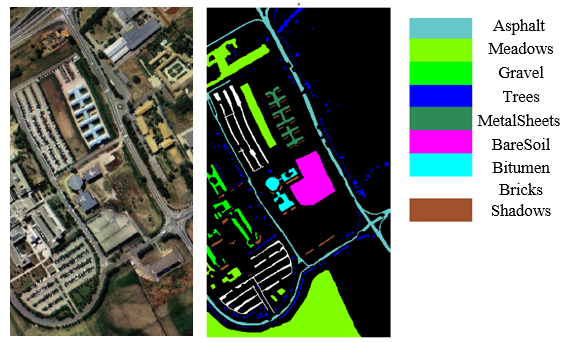} 
	\caption{The false-color composite image, the corresponding ground-truth map, and the legend of the Pavia University dataset.} 
	\label{fig:pavia_data} 
\end{figure}

The Salinas (SA) hyperspectral dataset was captured using an airborne sensor over Salinas Valley, California, USA. It consists of 512x217 pixels and 224 spectral bands ranging from 400 to 2500 nm, with a spatial resolution of about 3.7 meters. The dataset includes 16 crop types and has been widely utilized in classification, clustering, and feature extraction. Its availability has advanced hyperspectral imaging and agricultural monitoring research.

\begin{figure}[!h]
	\centering 
	\includegraphics[width=0.5\textwidth]{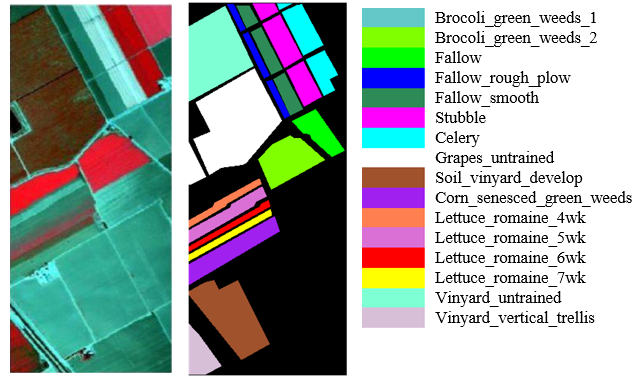} 
	\caption{The false-color composite image, the corresponding ground-truth map, and the legend of the Salinas dataset.} 
	\label{fig:salinas_data} 
\end{figure}

Table~\ref{tab:DataSet} shows the size of the training and testing datasets used in this experiment, including the sample distribution of each land cover category.

\subsubsection{Evaluation Metrics}
We will compare the effectiveness of our algorithm with other algorithms from four aspects, mainly including Overall Accuracy (OA), Average Accuracy (AA), kappa coefficient ($\kappa$), and the classification accuracy of each land cover category itself.
\subsubsection{Training Details}
The PyTorch training framework was utilized to implement and train the model, with the basic hardware environment comprising an AMD EPYC 7543 production-grade CPU, 128GB of memory, and two NVIDIA GeForce RTX 3090 GPUs, each with 24GB of memory. 

The diffusion model was optimized using the Adam optimizer, with a learning rate of 1e-4, a batch size of 256, and a patch size of 64. The selected patch size for the experiment is relatively large compared to the overall size of the remote sensing image. In this study, given the available computational resources, it is advisable to maximize the patch size in order to model a larger range of pixel relationships. In order to enhance the fitting performance of diffusion, it is recommended to use a large number of training epochs. For the IP, PU and SA datasets, this study utilized training results with more than 30000 epochs. The classification model was trained using an Adam optimizer, with a learning rate of 1e-3 and a smaller batch size of 64. Empirically, it was observed that this model converged rapidly, and fewer than 50 epochs were sufficient to achieve the desired classification accuracy for all three datasets. 

\subsection{Performance Analysis}

\begin{table*}[htpb]
\small
\begin{center}
    \caption{Classification Results of Experiments on the Indian Pines Dataset}

    \label{tab:TABLE_IP}
\begin{tabular}{c|c|c|c|c|c|c|c|c|c} \toprule Class& CNN1D \cite{9082155} & CNN2D \cite{2dcnn}& SF \cite{9627165}& miniGCN \cite{9170817}& SSRN \cite{8061020}& SSFTT \cite{tokenization}& DMVL\cite{multiview}& SSGRN\cite{ssgrn}& Ours\\  \midrule \midrule 1& 93.750& \textbf{100.000}& \textbf{100.000}& 12.500& \textbf{100.000}& \textbf{100.000}& 68.657& \textbf{100.000}& \textbf{100.000}\\ 2& 58.512& 63.877& 49.070& 65.880& \textbf{95.494}& 80.472& 92.023& 86.409& 85.479\\ 3& 57.500& 58.000& 71.500& 42.625& 84.875& 89.250& 89.074& 92.375& \textbf{94.125}\\ 4& 78.744& 79.227& 96.618& 60.870& 98.068& 99.034& 80.068& \textbf{100.000}& \textbf{100.000}\\ 5& 86.313& 80.795& 77.042& 84.989& 79.470& \textbf{99.338}& 96.742& 95.144& 96.026\\ 6& 90.571& 89.857& 85.857& 96.714& 93.429& 98.714& 85.697& 98.143& \textbf{99.857}\\ 7& 92.857& \textbf{100.000}& 85.714& 92.857& 92.857& \textbf{100.000}& 31.111& \textbf{100.000}& \textbf{100.000}\\ 8& 87.946& 94.643& 99.107& 98.214& \textbf{100.000}& \textbf{100.000}& 97.951& \textbf{100.000}& \textbf{100.000}\\ 9& 80.000& 70.000& \textbf{100.000}& 60.000& \textbf{100.000}& \textbf{100.000}& 74.074& \textbf{100.000}& \textbf{100.000}\\ 10& 54.246& 69.958& 80.149& 63.376& \textbf{96.178}& 91.720& 88.447& 86.837& 84.501\\ 11& 51.093& 46.887& 52.371& 37.237& 81.402& 88.454& \textbf{97.728}& 90.680& 91.959\\ 12& 69.805& 58.792& 52.753& 58.792& 91.652& 85.968& 84.301& 95.204& \textbf{95.560}\\ 13& 97.143& \textbf{100.000}& 98.857& 97.714& \textbf{100.000}& \textbf{100.000}& 93.488& \textbf{100.000}& \textbf{100.000}\\ 14& 76.923& 81.377& 79.109& 89.555& \textbf{99.757}& 99.028& 97.709& 92.793& 97.085\\ 15& 71.910& 63.483& 62.079& 53.652& 85.674& 98.596& 84.171& \textbf{100.000}& \textbf{100.000}\\ 16& 90.476& 88.889& \textbf{100.000}& 90.476& \textbf{100.000}& \textbf{100.000}& 39.732& \textbf{100.000}& 98.413\\  \midrule \midrule OA(\%)&66.007&66.966&67.782&63.916&90.658&91.566&90.487&92.336&\textbf{93.146}\\ AA(\%)&77.362&77.861&80.639&69.091&93.678&95.661&81.311&96.099&\textbf{96.438}\\ $\kappa \times 100 $&61.658&62.807&63.820&59.439&89.386&90.392&89.216&91.254&\textbf{92.175}\\  \bottomrule \end{tabular}

\end{center}
\end{table*}

\begin{table*}[]
\small
\begin{center}
    \caption{Classification Results of Experiments on the Pavia University Dataset}
\label{tab:TABLE_PU}
\begin{tabular}{c|c|c|c|c|c|c|c|c|c} \toprule Class& CNN1D \cite{9082155} & CNN2D \cite{2dcnn}& SF \cite{9627165}& miniGCN \cite{9170817}& SSRN \cite{8061020}& SSFTT \cite{tokenization}& DMVL\cite{multiview}& SSGRN\cite{ssgrn}& Ours\\  \midrule \midrule 1& 71.156& 66.444& 61.521& 60.627& \textbf{91.668}& 86.972& 76.361& 82.397& 86.154\\ 2& 85.273& 84.758& 87.975& 92.207& 91.664& 97.728& 96.995& 96.026& \textbf{98.394}\\ 3& 76.027& 73.659& 87.240& 79.894& 94.490& 92.943& 90.013& 93.717& \textbf{96.568}\\ 4& 77.192& 80.488& 94.100& 82.367& 91.595& 91.760& 66.530& \textbf{98.484}& 76.401\\ 5& 99.468& 99.163& \textbf{100.000}& 99.163& 98.327& 99.924& 86.293& 99.468& 99.544\\ 6& 88.138& 86.437& 65.573& 65.613& 99.480& 98.720& 97.113& \textbf{100.000}& 98.300\\ 7& 90.308& 86.462& 90.000& 96.154& 99.846& 99.000& 94.895& \textbf{99.923}& 99.077\\ 8& 79.107& 76.999& 78.614& 80.312& 96.166& 74.535& 97.243& 87.431& \textbf{98.987}\\ 9& \textbf{100.000}& 99.891& 99.455& \textbf{100.000}& 95.202& 99.455& 69.598& 92.475& 91.167\\  \midrule \midrule OA(\%)&82.772&81.424&81.511&82.355&93.636&93.667&89.768&93.850&\textbf{94.775}\\ AA(\%)&85.185&83.811&84.942&84.037&\textbf{95.382}&93.448&86.116&94.436&93.843\\ $\kappa \times 100 $&77.606&75.880&75.795&76.620&91.706&91.617&86.580&91.919&\textbf{93.061}\\  \bottomrule \end{tabular}

\end{center}

\end{table*}

\begin{table*}[htbp]
\small 
\begin{center}
\caption{Classification Results of Experiments on the Salinas Dataset}
\label{tab:TABLE_SA}
\begin{tabular}{c|c|c|c|c|c|c|c|c|c} \toprule Class& CNN1D \cite{9082155} & CNN2D \cite{2dcnn}& SF \cite{9627165}& miniGCN \cite{9170817}& SSRN \cite{8061020}& SSFTT \cite{tokenization}& DMVL\cite{multiview}& SSGRN\cite{ssgrn}& Ours\\  \midrule \midrule 1& 99.596& 99.747& 99.192& 99.141& \textbf{100.000}& \textbf{100.000}& 96.958& 92.976& \textbf{100.000}\\ 2& 99.729& 99.729& 99.513& 99.946& 95.996& \textbf{100.000}& 99.946& 95.860& \textbf{100.000}\\ 3& 97.431& 98.972& 84.275& 97.996& 81.449& \textbf{100.000}& 99.697& 98.561& \textbf{100.000}\\ 4& 99.267& 98.974& 98.754& 99.560& \textbf{100.000}& 99.927& 66.098& 93.109& \textbf{100.000}\\ 5& 97.432& 99.131& 92.674& 98.678& 91.843& 98.263& \textbf{99.875}& 95.582& 98.036\\ 6& 99.440& 99.211& 99.491& 99.822& 99.466& 99.949& 98.555& 99.746& \textbf{100.000}\\ 7& 99.606& 99.436& 99.127& 99.577& 99.944& 99.972& \textbf{100.000}& 98.929& 99.239\\ 8& 84.610& 65.599& 68.268& 77.493& 86.932& 82.929& 96.086& \textbf{97.545}& 96.975\\ 9& 98.753& 99.141& 97.732& 99.887& \textbf{100.000}& \textbf{100.000}& \textbf{100.000}& 99.417& \textbf{100.000}\\ 10& 89.070& 85.345& 88.208& 92.857& 96.121& 98.060& \textbf{100.000}& 88.762& 98.615\\ 11& 98.555& 90.944& 96.435& 95.857& \textbf{100.000}& \textbf{100.000}& 96.703& 97.399& 99.807\\ 12& 99.895& 99.736& 99.895& 99.789& \textbf{100.000}& \textbf{100.000}& 99.353& 99.051& 99.947\\ 13& 98.984& 98.307& 99.887& 98.984& \textbf{100.000}& 97.178& 94.731& 92.099& \textbf{100.000}\\ 14& 91.250& 96.058& 97.596& 96.250& 99.615& 99.519& 86.829& 99.231& \textbf{100.000}\\ 15& 64.424& 72.271& 81.044& 53.868& 90.564& 97.513& \textbf{99.090}& 93.396& 98.826\\ 16& 97.805& 93.697& 90.096& 99.043& \textbf{100.000}& 99.043& 99.724& \textbf{100.000}& \textbf{100.000}\\  \midrule \midrule OA(\%)&90.540&87.338&88.248&88.181&94.352&95.789&97.005&96.539&\textbf{98.971}\\ AA(\%)&94.740&93.519&93.262&94.297&96.371&98.272&95.853&96.354&\textbf{99.465}\\ $\kappa \times 100 $&89.451&85.931&86.973&86.823&93.722&95.322&96.668&96.144&\textbf{98.854}\\  \bottomrule \end{tabular}

\end{center}

\end{table*}

\begin{figure*}[htbp]
	\centering 
	\includegraphics[width=0.9\textwidth]{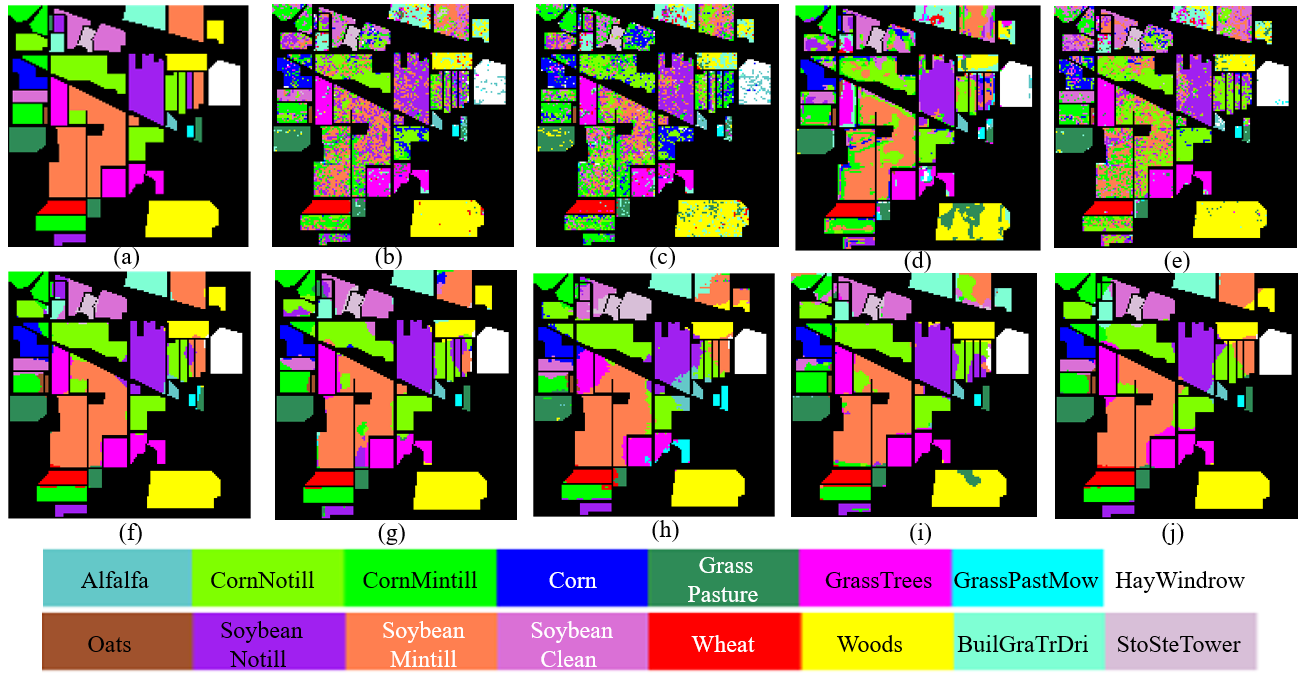} 
	\caption{Classification maps of the Indian Pines dataset. (a) Ground-truth map. (b) CNN1D (OA = 66.007\%). (c) CNN2D (OA = 66.966\%). (d) SF (OA = 67.782\%). (e) miniGCN (OA = 63.916\%). (f) SSRN (OA = 90.658\%). (g) SSFTT (OA = 91.566\%). (h) DMVL (OA = 90.487\%). (i) SSGRN (OA = 92.336\%). (j) Ours (OA = 93.146\%).} 
	\label{fig:res_indian} 
	
\end{figure*}

\begin{figure*}[htbp]
	\centering 
	\includegraphics[width=0.9\textwidth]{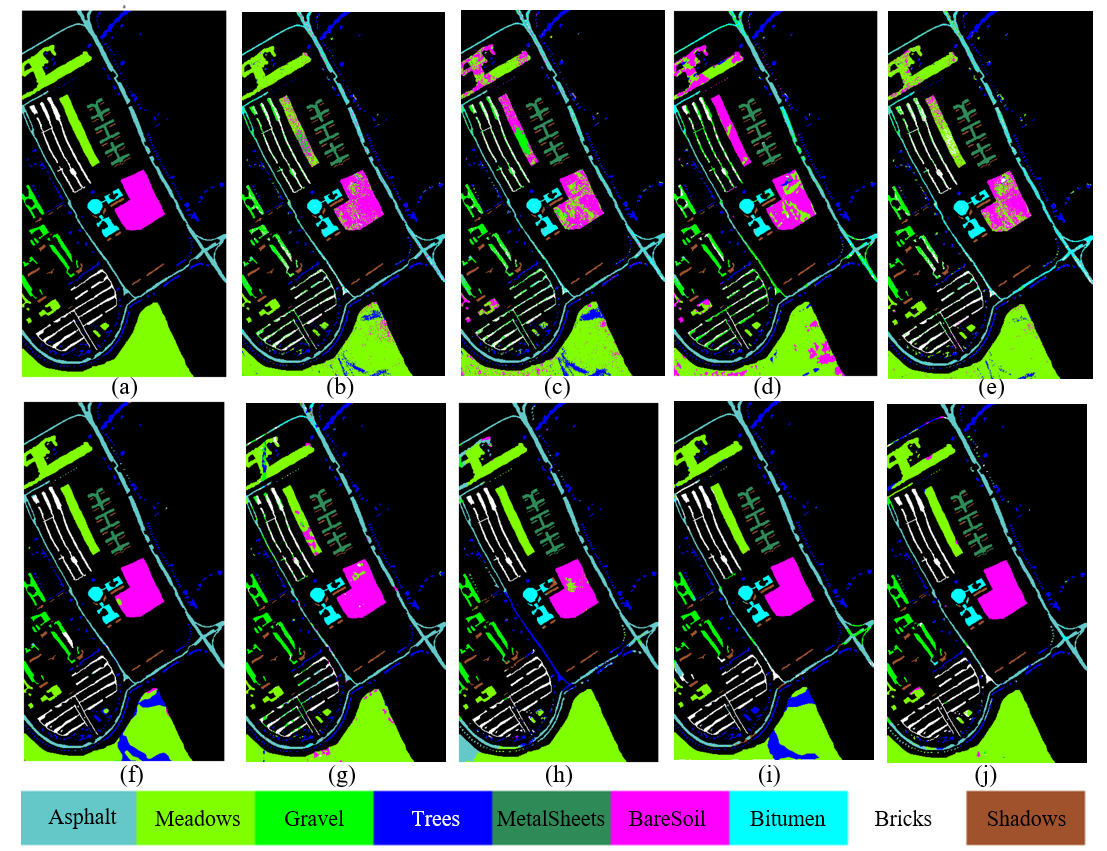} 
	\caption{Classification maps of the Pavia University dataset. (a) Ground-truth map. (b) CNN1D (OA = 82.772\%). (c) CNN2D (OA = 81.424\%). (d) SF (OA = 81.511\%). (e) miniGCN (OA = 82.355\%). (f) SSRN (OA = 93.636\%). (g) SSFTT (OA = 93.667\%). (h) DMVL (OA = 89.768\%). (i) SSGRN (OA = 93.850\%). (j) Ours (OA = 94.775\%).
} 
	\label{fig:res_pavia} 
\end{figure*}

\begin{figure*}[htbp]
	\centering 
	\includegraphics[width=0.9\textwidth]{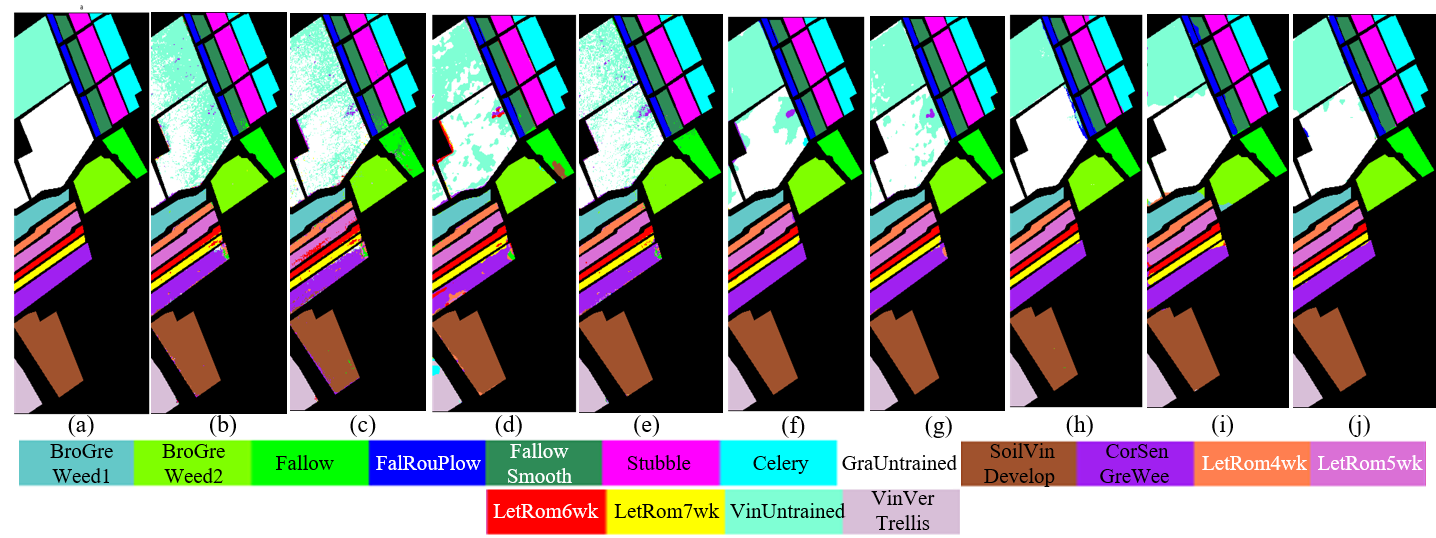} 
	\caption{Classification maps of the Salinas dataset. (a) Ground-truth map. (b) CNN1D (OA = 90.540\%). (c) CNN2D (OA = 87.338\%). (d) SF (OA = 88.248\%). (e) miniGCN (OA = 88.181\%). (f) SSRN (OA = 94.352\%). (g) SSFTT (OA = 95.789\%). (h) DMVL (OA = 97.005\%). (i) SSGRN (OA = 96.539\%). (j) Ours (OA = 98.971\%).} 
	\label{fig:res_salinas} 
\end{figure*}

\begin{figure*}[htbp]
	\centering 
	\includegraphics[width=0.9\textwidth]{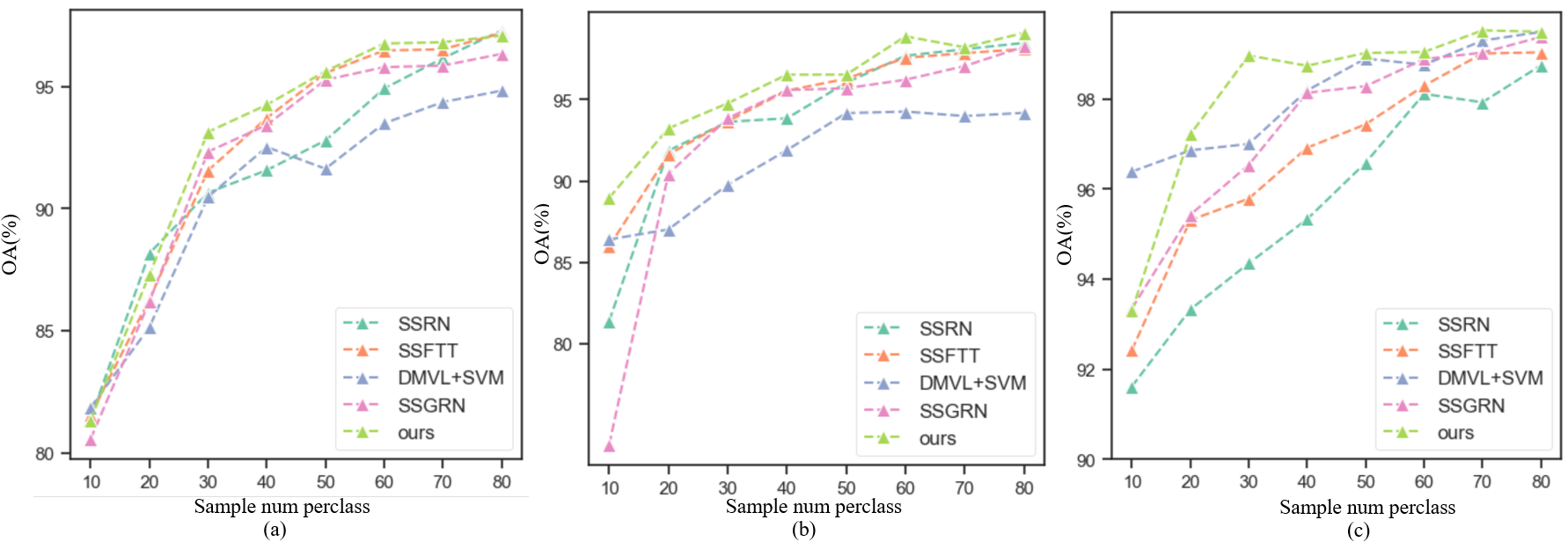} 
	\caption{Evolution of OA as a function of number of training samples per class. (a) Indian Pines. (b) University of Pavia.(c) Salinas.} 
	\label{fig:res_oa_persample} 
\end{figure*}

To assess the effectiveness of the proposed approach, various representative algorithms, namely CNN1D, CNN2D, SF, miniGCN, SSRN, SSFTT, DMVL(+SVM), and SSGRN, were selected for the control experiments.
\begin{enumerate}

	\item[1)]CNN1D\cite{9082155}, consists of 5 layers: a 1-D convolutional layer, a batch normalization (BN) layer, a rectified linear unit (ReLU) layer, an average pooling layer, and an output layer. CNN1D is one of the classical CNN algorithms, and our proposed algorithm also includes a CNN structure.
	
	\item[2)]CNN2D\cite{2dcnn}, consists of three 2-D convolutional blocks, each of which contains a 2-D convolutional layer, a BN layer, and a ReLU activation function. Each 2-D convolutional block has 8, 16, and 32 $3\times3$ 2-D filters, respectively. CNN2D is one of the classical CNN algorithms, and our proposed algorithm also includes a CNN structure.
	
	\item[3)]SF\cite{9627165}, SpectralFormer is one of the classical transformer-based algorithms, and our structure also incorporates a transformer structure.
	
	\item[4)]miniGCN\cite{9170817}, a variant of graph convolutional networks (GCN) designed to train large-scale GCNs in a mini-batch fashion, making it more efficient and flexible than traditional GCNs. miniGCN is one of the classical algorithms for constructing sample relationships.
	
	\item[5)]SSRN\cite{8061020}, Spectral-Spatial Residual Network, enhances the effectiveness of CNN-based models by fusing and extracting spectral-spatial information through the Spectral-Spatial Residual Network. SSRN represents one of the state-of-the-art CNN-based methods.
	
	\item[6)]SSFTT\cite{tokenization}, Spectral-Spatial Feature Tokenization Transformer, effectively integrates CNN and Transformers, representing high-level semantic features, and achieving state-of-the-art performance.
	
	\item[7)]DMVL(+SVM)\cite{multiview}, Deep Multiview Learning(+SVM) performs unsupervised feature extraction followed by classification using an SVM classifier. Similar to our proposed algorithm, DMVL belongs to the two-stage algorithms, and its classification performance reaches one of the state-of-the-art levels.
	
	\item[8)]SSGRN\cite{ssgrn}, Spectral-Spatial Graph Reasoning Network, employs intermediate features to generate superpixels, facilitating the creation of homogeneous regions and graph structures. SSGRN is one of the state-of-the-art graph-based algorithms. 
	
	\item[9)] Ours: Our proposed model follows the basic architecture described earlier, comprising two stages: diffusion model training and classification. During diffusion model training, the input data is divided into $64\times 64$ patches, which are first passed through a 3D convolution layer with a kernel size of (3,3,3) and padding size of (1,1,1). Subsequently, three 3D down-sampling and three 3D up-sampling layers are employed for further spatial and spectral feature extraction. Finally, another 3D convolution layer with a kernel size of (3,3,3) and padding size of (1,1,1) will be passed. To extract diffusion features, we performed spatial and spectral reconstruction with noise images and timestamps as inputs. We used the output of U-net down-sampling and up-sampling layers while reducing its dimensionality to serve as diffusion features. In this study, PCA was used. In the classification stage, the diffusion features go through a series of layers, including the Convolutional Tokenization layer, Linear Projection layer, transformer encoders, and MLP layer, to ultimately obtain the classification results. At this stage, the patch size is set to 13.
\end{enumerate}

\subsubsection{Quantitative Results}
Tables~\ref{tab:TABLE_IP}-\ref{tab:TABLE_SA} respectively presents the classification performance of the IP, PU and SA datasets, including OA, AA, $\kappa$, and the classification accuracy of each category. The best performance is highlighted in bold. It is worth noting that our model exhibits the best overall performance on these three datasets.

Through detailed analysis, it can be observed from the table that traditional algorithms such as CNN1D and CNN2D exhibit relatively poor performance in classification. This is primarily because these models have weak feature extraction capabilities. Especially for HSIs, simple models struggle to effectively capture both spatial and spectral features, particularly when dealing with abundant spectral data. On the other hand, the SpectralFormer and miniGCN models are more complex, but their performance is relatively inferior in limited sample situations.

The SSRN, SSFTT, DMVL and SSGRN algorithms perform relatively well. The SSRN and SSFTT models deeply extract spectral-spatial information, leading to a significant improvement in classification performance. The DMVL algorithm effectively utilizes unsupervised information. Additionally, the SSGRN algorithm enhances model performance through additional modeling of global relationships. Our proposed algorithm enhances the modeling of the spectral-spatial relationship through the diffusion process, surpassing the aforementioned algorithms. Furthermore, it integrates spectral-spatial features more effectively using the transformer architecture, resulting in superior performance. In comparison to the second-best algorithm, our approach achieved an improvement of $0.81\%$, $0.93\%$, and $1.97\%$ in overall accuracy (OA) on the IP, PU and SA datasets, respectively.

From an analysis of the classification performance in different subcategories, it can be concluded that our algorithm demonstrates relatively balanced results across various subcategories in the IP and SA datasets, with higher values of AA and $\kappa$. However, on the PU dataset, our algorithm exhibits relatively weaker performance in achieving balanced classification results among different subcategories compared to models such as SSRN, Nevertheless, the overall AA remains high.

Fig.~\ref{fig:res_oa_persample} presents a comprehensive comparison of the overall accuracy among corresponding models using the IP, PU, and SA datasets with varying sample sizes. The figure clearly demonstrates that increasing the sample size leads to a continuous improvement in the overall classification accuracy for each model. Remarkably, our proposed model consistently outperforms other models across a wide range of sample sizes. When the number of samples per-class drops to 10, our model still demonstrates the best performance on the PU dataset, while it does not exhibit significant advantages on the IP and SA datasets. We consider this may be attributed to the larger number of feature channels extracted through the diffusion process, resulting in a certain degree of overfitting in the classification model.

\subsubsection{Qualitative results}
Figs.~\ref{fig:res_indian}-\ref{fig:res_salinas} display the classification results of IP, PU and SA datasets. Visually, our model exhibits lower noise levels and closely aligns with the Ground Truth. Traditional classification algorithms such as CNN1D, CNN2D, SF and miniGCN produce noisy classification maps with discontinuous land cover blocks and rough classification results. Meanwhile, SSRN, SSFTT, DMVL and SSGRN algorithms show improved performance, with a reduction in noisy points. Furthermore, from a detailed perspective, our proposed algorithm exhibits better overall segmentation of land cover.

\subsection{Model Analysis}

\subsubsection{Ablation study}
\begin{figure*}[htbp]
	\centering 
	\includegraphics[width=0.9\textwidth]{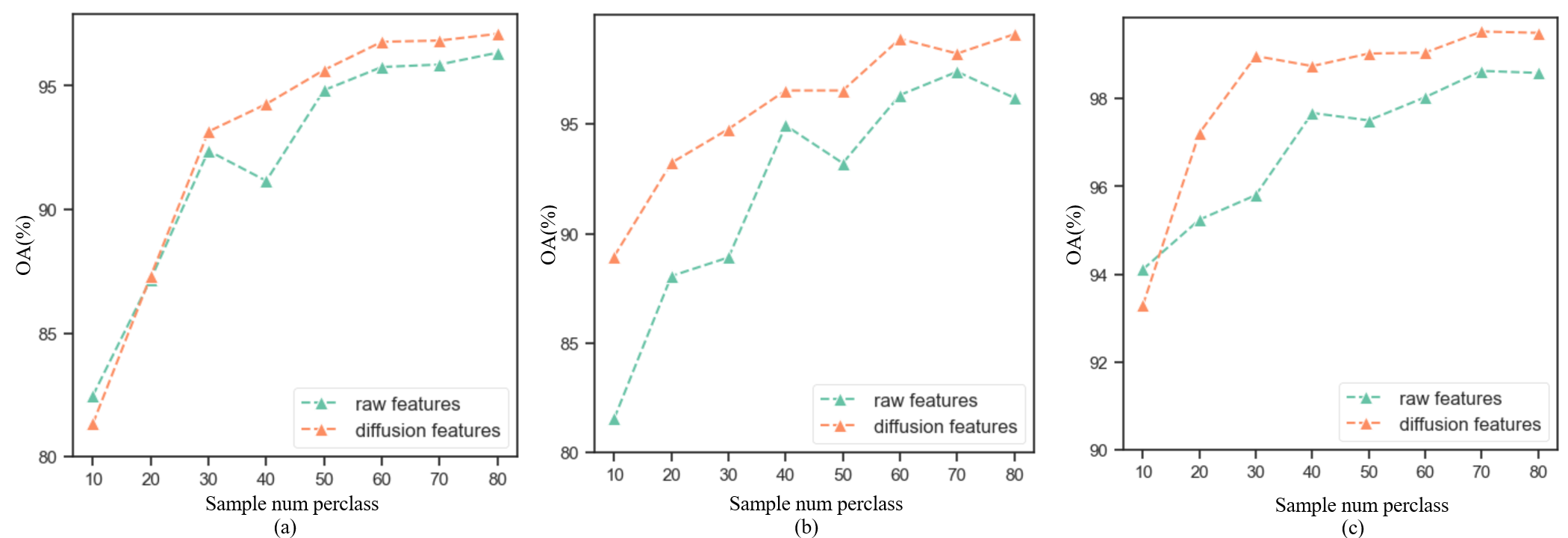} 
	\caption{Overall Accuracy with Varying Sample Sizes: Raw Features Inputs vs. Diffusion Features Inputs. (a) Indian Pines. (b) University of Pavia.(c) Salinas.} 
	\label{fig:ab_diff} 
\end{figure*}

 We further analyzed the benefits of the proposed SpectraiDiff, which involves feeding the original spectral features and the features extracted from the pre-training process with the diffusion model into our proposed attention-based classification model. As shown in Table~\ref{tab:DIF}, the classification results on IP, PU and SA datasets are presented. The results demonstrate that using diffusion features as input significantly outperforms the use of raw features on all three datasets, resulting in improved OA, AA, and $\kappa$ metrics. It is worth noting that using diffusion features as input, compared to using raw features as input, results in an overall classification accuracy improvement (OA) of $0.78\%$, $5.87\%$, and $3.17\%$ on the IP, PU, and SA datasets, respectively. Fig.~\ref{fig:ab_diff} demonstrates the improvement in performance under different sample sizes. Within the framework of the proposed Attention-based Classification Module, the variation in the number of training samples per class from 20 to 80 reveals a consistent trend: OA achieved with diffusion features as input consistently surpasses that obtained with raw features. However, when the sample size decreases to 10 per class, the inclusion of diffusion features does not yield any performance improvement on both the IP and PU datasets. This occurrence can be attributed to the potential overfitting of the classification model, which arises due to the high dimensionality of the generated diffusion features. 
\begin{table}[h]
	\small
	\begin{center}
		\normalsize
		\caption{\textsc{Classification Performance Analysis Between Different Feature Input On The Indian Pines Dataset, Pavia University Dataset And Salinas Dataset}}
		\label{tab:DIF}
		\setlength{\tabcolsep}{1.5mm}
		\begin{tabular}{c|c|ccc}
			\toprule [1.5pt]
			
			\multicolumn{1}{c|}{\multirow{2}{*}{Dataset}} & 
			\multicolumn{1}{c|}{\multirow{2}{*}{Feature Input}}   &
			\multicolumn{3}{c}{Metrics} \\\cline{3-5} 
			
			\multicolumn{1}{c|}{}  & 
			\multicolumn{1}{c|}{}  & 
			\multicolumn{3}{c}{  OA(\%)   AA(\%)   $\kappa$*100} \\
			
			\midrule
			\midrule
			\multirow{2}{*}{IP}
                    & raw features &92.37 & 95.61 & 91.27\\
    			& diffusion features &93.15 & 96.44 & 92.17\\ \cline{1-1} \cline{2-5}
			
			\multirow{2}{*}{PU} & raw features &88.91 & 89.77 & 85.51\\
			&  diffusion features &94.78 & 93.84 & 93.06\\  \cline{1-1} \cline{2-5}
			
			\multirow{2}{*}{SA} & raw features &95.80 & 98.26 & 95.34\\
			& diffusion features &98.97 & 99.47 & 98.85\\
			
			\bottomrule [1.5pt]
		\end{tabular}
	\end{center}
\end{table}

\subsubsection{Diffusion Model Analysis}

	

\begin{table*}[htbp]
	\small 
	\begin{center}
		\caption{\textsc{The Performance Of Different Layerindex And Timestamp In The Indian Pines Dataset, The Pavia University Dataset, And Salinas Dataset}}
		\setlength{\tabcolsep}{2mm}
		\label{tab:res_layerindex_timestamp}
		
		\begin{tabular}{c|c||ccc||ccc||ccc}
			\toprule[1.5pt]
			
			\multicolumn{1}{c|}{\multirow{2}{*}{LayerIndex}} & 
			\multicolumn{1}{c||}{\multirow{2}{*}{Timestamp}} & 
			\multicolumn{3}{c||}{\textbf{Indian Pines}} & 
			\multicolumn{3}{c||}{\textbf{Pavia University}}  & 
			\multicolumn{3}{c}{\textbf{Salinas}}  \\           
			
			\multicolumn{1}{c|}{}  & 
			\multicolumn{1}{c||}{}  & 
			\multicolumn{1}{c}{OA(\%)} & \multicolumn{1}{c}{AA(\%)} & \multicolumn{1}{c||}{$\kappa*100$} &
			\multicolumn{1}{c}{OA(\%)} & \multicolumn{1}{c}{AA(\%)} & \multicolumn{1}{c||}{$\kappa*100$} &
			\multicolumn{1}{c}{OA(\%)} & \multicolumn{1}{c}{AA(\%)} & \multicolumn{1}{c}{$\kappa*100$} \\
			
			\midrule
			\midrule
			 {\multirow{5}{*}{0}} & 5 & 92.73 & 95.96 & 91.69 & \textbf{94.78} & 93.84 & \textbf{93.06} & \textbf{98.97} & \textbf{99.47} & \textbf{98.85} \\
              & 10 & 92.19 & 96.08 & 91.08 & 93.60 & 93.14 & 91.57 & 98.64 & 99.35 & 98.49 \\
              & 50 & 92.67 & 96.10 & 91.62 & 93.48 & 91.79 & 91.37 & 97.73 & 99.05 & 97.47 \\
              & 100 & 92.43 & 95.93 & 91.35 & 92.86 & 92.04 & 90.53 & 97.80 & 98.87 & 97.56 \\
              & 200 & 91.39 & 95.56 & 90.18 & 91.62 & 90.84 & 88.96 & 97.70 & 98.72 & 97.44 \\
             \midrule 
             {\multirow{5}{*}{1}} & 5 & 92.97 & 96.43 & 91.98 & 94.31 & \textbf{94.19} & 92.53 & 98.69 & 99.22 & 98.54 \\
              & 10 & 92.41 & 96.11 & 91.34 & 94.72 & 93.83 & 93.00 & 98.32 & 99.20 & 98.13 \\
              & 50 & 91.90 & 95.92 & 90.75 & 92.73 & 93.03 & 90.46 & 97.99 & 98.94 & 97.76 \\
              & 100 & 92.12 & 95.80 & 90.99 & 94.12 & 92.79 & 92.19 & 97.98 & 99.08 & 97.75 \\
              & 200 & 92.02 & 95.89 & 90.89 & 91.84 & 90.55 & 89.27 & 96.73 & 98.42 & 96.36 \\
             \midrule 
             {\multirow{5}{*}{2}} & 5 & 93.15 & \textbf{96.44} & \textbf{92.17} & 92.02 & 91.65 & 89.52 & 98.02 & 99.03 & 97.80 \\
              & 10 & 92.85 & 96.39 & 91.84 & 91.26 & 91.02 & 88.55 & 97.74 & 98.77 & 97.49 \\
              & 50 & 92.93 & 96.21 & 91.93 & 91.10 & 90.66 & 88.23 & 95.51 & 97.62 & 95.01 \\
              & 100 & \textbf{93.16} & 96.32 & \textbf{92.17} & 86.51 & 85.12 & 82.25 & 95.96 & 97.47 & 95.51 \\
              & 200 & 92.60 & 96.11 & 91.55 & 81.54 & 81.72 & 75.88 & 93.53 & 96.37 & 92.80 \\
			
			\bottomrule[1.5pt]
		\end{tabular}
	
	\begin{tablenotes}
		\footnotesize
		\item{1} LayerIndex=0, 1, 2 respectively represent the inputs of the three up-sampling layers in the U-Net model, with larger numbers closer to the output layer.
		\item{2} A smaller Timestamp indicates that the diffusion denoising process is closer to the original image position.
	\end{tablenotes}
	
	\end{center}
\end{table*}
\begin{figure*}[htbp]
	\centering 
	\includegraphics[width=0.9\textwidth]{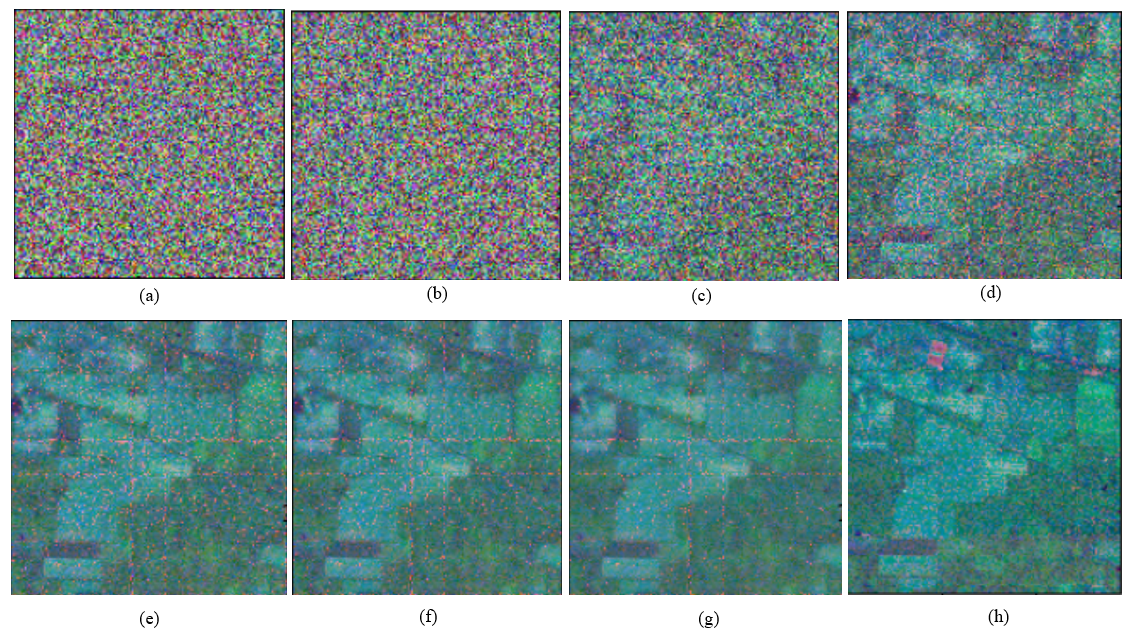} 
	\caption{False-color images of the reconstructed IndianPines dataset corresponding to different timestamps by Reverse Spectral-Spatial Diffusion Process. (a) $t=400$. (b) $t=200$. (c) $t=100$. (d) $t=50$. (e) $t=10$. (f) $t=5$. (g) $t=0$. (h) ground truth. } 
	\label{fig:diffusion} 
	
\end{figure*}

\begin{figure*}[htbp]
	\centering 
	\includegraphics[width=0.9\textwidth]{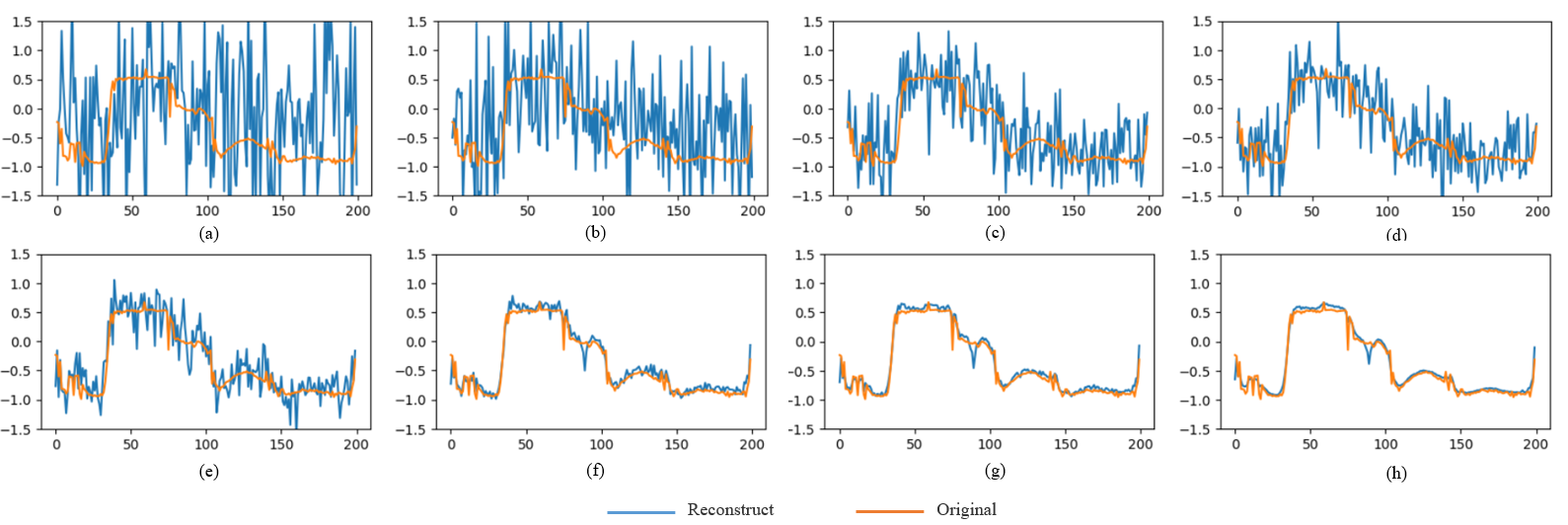} 
	\caption{The spectral curves of the reconstructed Indian Pines dataset corresponding to different timestamps by Reverse Spectral-Spatial Diffusion Process, class=woods, x-axis:spectral band number, y-axis: normalized spectral values. (a) $t=400$. (b) $t=200$. (c) $t=100$. (d) $t=80$. (e) $t=50$. (f) $t=10$. (g) $t=5$. (h) $t=0$.} 
	\label{fig:diffusion_curve1} 
	
\end{figure*}
\begin{table}[h]
	\small
	\begin{center}
		\normalsize
		\caption{\textsc{GPU Inference Time (s) of the Proposed Method and the Comparison Methods}}
		\label{tab:time}
		\setlength{\tabcolsep}{1.5mm}
		\begin{tabular}{c||c|c|c}
			\toprule[1.5pt]
			
			\multicolumn{1}{c||}{Method} & 
                \multicolumn{1}{c|}{\textbf{IP}} & 
                \multicolumn{1}{c|}{\textbf{PU}} & 
                \multicolumn{1}{c}{\textbf{SA}}    \\           
			
			\midrule
			\midrule
			CNN1D & 9.74 & 37.94 & 20.01 \\
                CNN2D & 14.65 & 45.84 & 24.74 \\
                SF & 29.81 & 163.34 & 141.30 \\
                miniGCN & 0.95 & 2.40 & 2.13 \\
                SSRN & 48.02 & 248.06 & 222.80 \\
                SSFTT & 31.58 & 110.08 & 55.96 \\
                SSGRN & 1.13 & 4.99 & 2.98 \\
                Ours & 48.53 & 256.23 & 242.16 \\
			
			\bottomrule[1.5pt]
		\end{tabular}
	\end{center}
\end{table}

We further analyzed the effects of the diffusion model on the final classification model. Firstly, to verify the diffusion model effectiveness, we used the well-trained diffusion model to recover and reconstruct the spectral curves of the hyperspectral data. we utilized HIS images corrupted with Gaussian noise as the input, and performed the Reverse Spectral-Spatial Diffusion Process step by step for each timestamp. As an example, Fig.~\ref{fig:diffusion} presents the restoration effects corresponding to different timestamps on the Indian Pines dataset using a false-color image. From a visual perspective, the diffusion model basically reconstructs the original remote sensing image content. Additionally, Fig.~\ref{fig:diffusion_curve1} shows the reconstruction process for spectral curve of one land-cover type. As the timestamps change, the spectral curve gradually reconstructs to the original shape of the land-cover types from a state similar to white noise. This indicates that the diffusion model has embedded the spectral curve information into the model parameters, providing a data foundation for using the diffusion feature for land-cover classification.

When extracting features using the Diffusion model, there are two crucial influencing parameters to consider, namely the Timestamp and the Layerindex. Timestamp refers to the number of denoising steps the Diffusion model takes to restore noisy images additionally. Layerindex refers to the location of the U-Net output used as a feature layer in the Diffusion model. We have conducted classification experiments on various Timestamp and Layerindex values, and the results are presented in the Table~\ref{tab:res_layerindex_timestamp}. In the case of the IP dataset, there are some fluctuations in classification performance for different Timestamp and Layerindex values, but no significant changes. We postulate that this may be attributed to the relatively small size of the IP dataset. However, for the PU and SA datasets, there is a certain correlation between classification performance and Timestamp/Layerindex. When considering the Timestamp dimension, a decreasing trend in classification performance is observed when using features with larger Timestamp, and the optimal performance generally occurs in smaller Timestamp groups (Timestamp = 5). We believe that when the Timestamp is larger, the number of iterations for denoising using the Diffusion model is lower, leading to relatively more noise information in the input and resulting in a deviation in the classification performance. Considering the Layerindex dimension, both datasets (PU \& SA) showed better performance at Layerindex 0 than at Layerindex 1 and 2.

We compared the inference time between different algorithms. It is noteworthy that our algorithm, being a two-stage algorithm, only accounts for the inference time at the classification stage in this analysis. Table.~\ref{tab:time} shows the testing time for each algorithm. Our algorithm takes longer compared to simple CNN algorithms such as CNN1D, CNN2D, and GCN algorithms. However, the increase in time is not significant when compared to complex CNN algorithms like SSRN and transformer algorithms, as they are still at the same level.

\section{Conclusion}
In this study, a novel approach is proposed for constructing the spectral-spatial distribution of HSI data from a generative perspective and capturing the spectral-spatial features. The proposed method provides a unique viewpoint for the spectral-spatial diffusion process and plays a critical role in establishing relationships between samples. With the proposed SpectralDiff, sample relationships can be adaptively constructed without prior knowledge of graph structure or neighborhood information. This approach captures the data distribution and contextual information of objects in HSI, achieving cross-sample perception. Experimental results demonstrate that this method outperforms state-of-the-art techniques.

Looking towards the future, an exciting avenue of exploration lies in studying the potential of diffusion models for out-of-distribution generalization and detection in the context of hyperspectral imaging, building upon the generative paradigm. It is expected that diffusion models will continue to advance these areas, capturing underlying data manifolds through diffusion processes, thereby learning to generalize well to unseen examples lying outside the training distribution. Additionally, diffusion models demonstrate strong detection performance in identifying out-of-distribution samples. Going forward, there is significant potential for diffusion models to further contribute to the fields of out-of-distribution generalization and detection. With further research, developments in leveraging the power of diffusion models to analyze complex and high-dimensional hyperspectral data are expected to continue, leading to exciting opportunities for future applications in diverse areas.


%

\FloatBarrier

\ifCLASSOPTIONcaptionsoff
  \newpage
\fi



%
\bibliographystyle{IEEEtran}
\bibliography{SpectralDiff_LaTeX}

%

\vspace{-10 mm}
\begin{IEEEbiography}[{\includegraphics[width=1in,height=1.25in,clip,keepaspectratio]{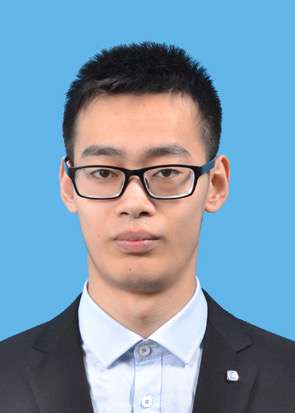}}]{Ning Chen}
	received the B.S. degree from the School of Earth and Space Sciences, Peking University, Beijing, China, in 2016, and the M.S. degree in GIS from the School of Earth and Space Sciences, Peking University, Beijing, China in 2019.
	
	He is currently an Assistant Engineer with the Institute of Remote Sensing and Geographic Information System, Peking University. His research interests include satellite image understanding, recommendation system, large-scale sparse learning and pattern recognition.
\end{IEEEbiography}
\vspace{-0 mm}
\begin{IEEEbiography}[{\includegraphics[width=1in,height=1.25in,clip,keepaspectratio]{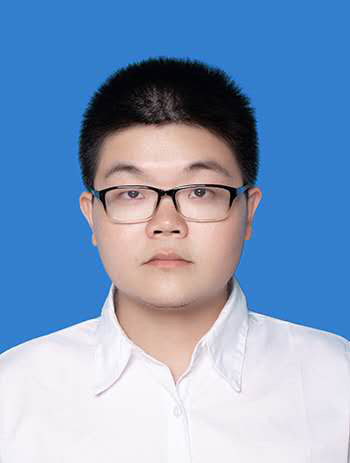}}]{Jun Yue}
	received the B.Eng. degree in geodesy from Wuhan University, Wuhan, China, in 2013 and the Ph.D. degree in GIS from Peking University, Beijing, China, in 2018. 
	
	He is currently an Assistant Professor with the School of Automation, Central South University. His research interests include satellite image understanding, pattern recognition, and few-shot learning. Dr. Yue serves as a reviewer for IEEE Transactions on Image Processing, IEEE Transactions on Neural Networks and Learning Systems, IEEE Transactions on Geoscience and Remote Sensing, ISPRS Journal of Photogrammetry and Remote Sensing, IEEE Geoscience and Remote Sensing Letters, IEEE Transactions on Biomedical Engineering, Information Fusion, Information Sciences, etc.
	
\end{IEEEbiography}
\vspace{-0 mm}
\begin{IEEEbiography}[{\includegraphics[width=1in,height=1.25in,clip,keepaspectratio]{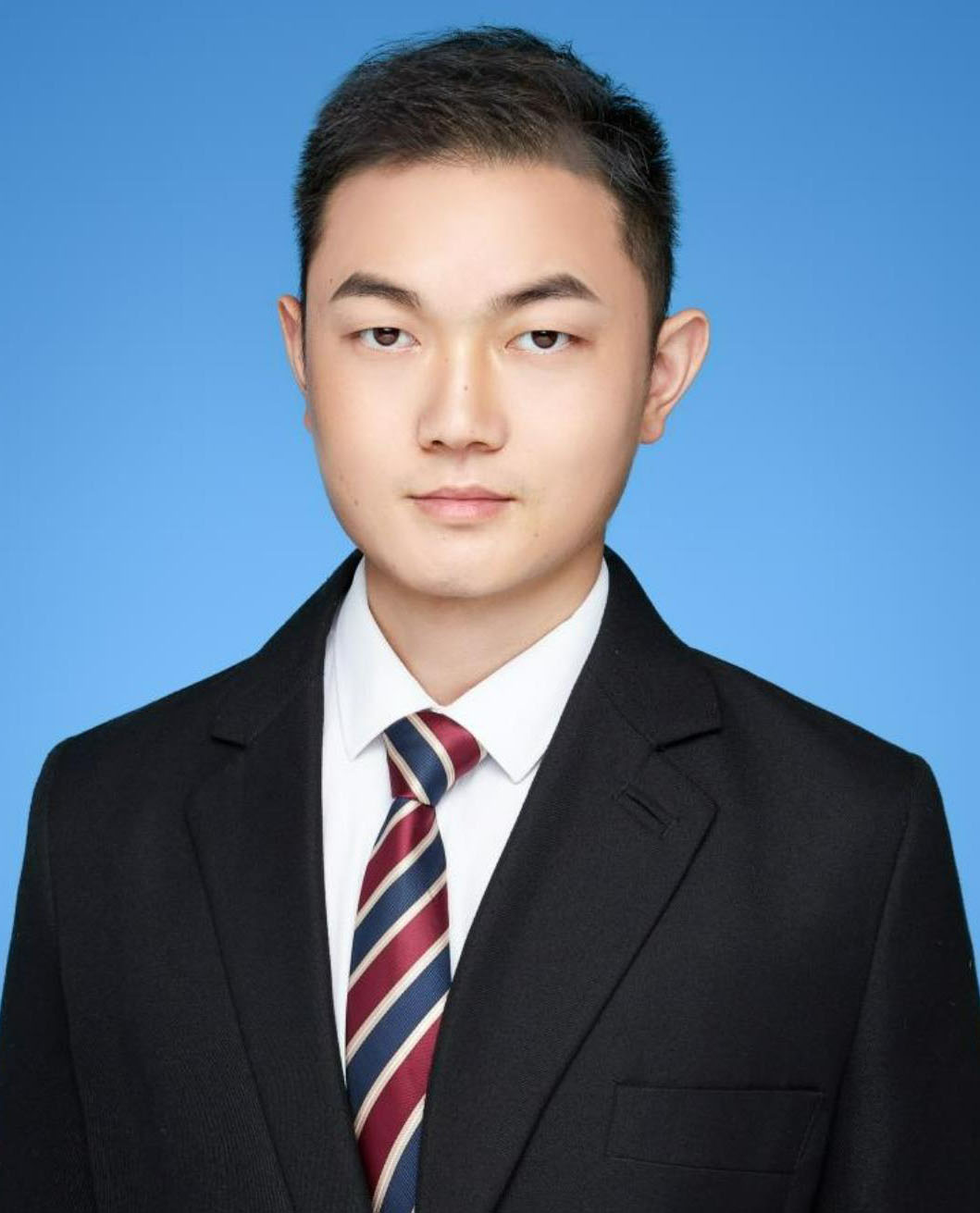}}]{Leyuan Fang}
	(Senior Member, IEEE) received the Ph.D. degree from the College of Electrical and Information Engineering, Hunan University, Changsha, China, in 2015. 
	
	From August 2016 to September 2017, he was a Postdoc Researcher with the Department of Biomedical Engineering, Duke University, Durham, NC, USA. He is currently a Professor with the College of Electrical and Information Engineering, Hunan University. His research interests include sparse representation and multi-resolution analysis in remote sensing and medical image processing. He is the associate editors of IEEE Transactions on Image Processing, IEEE Transactions on Geoscience and Remote Sensing, IEEE Transactions on Neural Networks and Learning Systems, and Neurocomputing. He was a recipient of one 2nd-Grade National Award at the Nature and Science Progress of China in 2019.  
\end{IEEEbiography}
\vspace{-80 mm}
\begin{IEEEbiography}[{\includegraphics[width=1in,height=1.25in,clip,keepaspectratio]{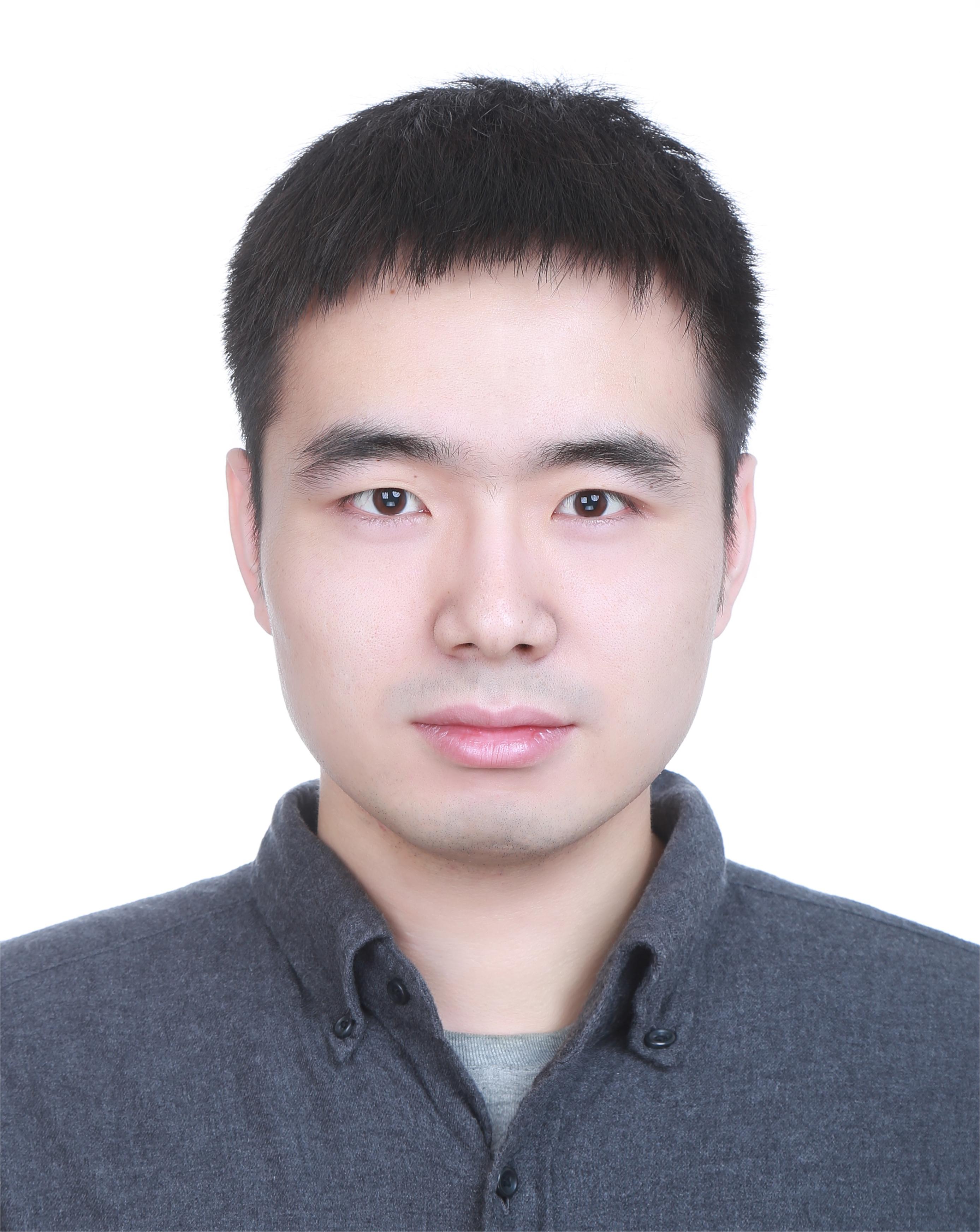}}]{Shaobo Xia} received the bachelor's degree in geodesy and geomatics from the School of Geodesy and Geomatics, Wuhan University, Wuhan, China, in 2013, the master's degree in cartography and geographic information systems from the Institute of Remote Sensing and Digital Earth, Chinese Academy of Sciences, Beijing, China, in 2016, and the Ph.D. degree in geomatics from the University of Calgary, Calgary, AB, Canada, in 2020.
	
	He is an Assistant Professor with the Department of Geomatics Engineering, Changsha University of Science and Technology, Changsha, China. His research interests include point cloud processing and remote sensing.
\end{IEEEbiography}





\FloatBarrier
\end{document}